\documentclass[12pt]{article}

\usepackage[utf8]{inputenc}
\usepackage[T1]{fontenc}
\usepackage{amsmath}  %
\usepackage{txfonts}  %
\usepackage[top=25mm,bottom=25mm,left=30mm,right=30mm]{geometry}
\usepackage{graphicx}
\graphicspath{{figures/}{figures/supplementary/}}  %
\usepackage{amssymb,amsfonts}
\usepackage{caption}
\usepackage[subrefformat=parens]{subcaption}
\usepackage{booktabs}
\usepackage{multirow}
\usepackage{here}
\usepackage{subcaption} %
\usepackage{xcolor}     %
\usepackage{longtable}  %
\usepackage{enumitem}   %
\usepackage[section]{placeins} %

\newif\ifshowlineno    \showlinenofalse
\newif\ifshowrevisions \showrevisionsfalse

\ifshowlineno
  \usepackage{lineno}
  \linenumbers
\fi
\ifshowrevisions\else
  \colorlet{red}{black} %
\fi

\usepackage[backend=biber,style=vancouver,url=false,doi=false,isbn=false, sorting=none,sortcites=true,giveninits=true,maxbibnames=3,minbibnames=3,language=auto]{biblatex}



\let\cite\supercite

\title{\textbf{Anthropomimetic Soft Robotic {\color{red}Forearm} with Independently Articulated Carpal Bones {\color{red}Enabling} Human-Like Adaptive Stiffness Modulability}}

\author{
Yoshinobu Obata$^{1}$, %
Yinlai Jiang$^{2}$, %
Hiroshi Yokoi$^{1, 2}$, %
Shunta Togo$^{1, 2}$%
\\[1ex]
\small $^{1}$Department of Mechanical and Intelligent Systems Engineering,\\
\small The University of Electro-Communications, Tokyo, Japan\\
\small $^{2}$Center for Neuroscience and Biomedical Engineering,\\
\small The University of Electro-Communications, Tokyo, Japan\\
}

\date{}

\begin{document}

\maketitle

\begin{abstract}
\noindent
The human wrist exhibits adaptive stiffness modulability: joint stiffness anisotropy can be actively regulated through muscle co-contraction. This functionality is essential for stable manipulation, yet the underlying morphological factors remain unclear. To identify these factors, we developed an anatomically accurate anthropomimetic soft robotic {\color{red}forearm} comprising eight independently movable carpal bones interconnected by ligaments, 22 actuated muscles, and compliant fingertips. We measured wrist joint stiffness under four muscle activation patterns across three skeletal configurations: anatomically normal carpal bones, a fused proximal carpal row, and a {\color{red}geometric} ellipsoidal skeleton. The stiffness ellipse exhibited low stiffness along the dart-throwing motion (DTM) direction when finger muscles were activated, but high stiffness along the same direction when wrist and finger muscles were activated simultaneously. These results agree with previously reported human measurements, demonstrating that precise anatomical replication reproduces human-like stiffness modulability. Fusing the proximal carpal row eliminated the low DTM-direction stiffness under finger muscle activation, while the {\color{red}geometric} ellipsoidal skeleton showed poor stiffness ellipse reorientation across all conditions. Carpal bone motion analysis revealed significantly opposing coupling patterns between wrist and finger muscles at the proximal carpal row, accompanied by a consistent but non-significant trend at the midcarpal joint, providing a mechanical explanation for this modulation. These findings demonstrate that carpal bone morphology plays a dominant role in human wrist stiffness modulation and provide design principles for humanoid robot wrists.

\end{abstract}

\medskip
\noindent\textbf{Keywords:} soft robotics; biomimetic robot; wrist stiffness; carpal bones; anthropomimetic; adaptive stiffness; dart-throwing motion

\newpage

{\color{red}
\section*{List of Abbreviations}
\label{sec:abbreviations}

\begin{table}[h]
\centering
\begin{tabular}{ll}
\toprule
Abbreviation & Definition \\
\midrule
AP    & Adductor pollicis               \\
APB   & Abductor pollicis brevis        \\
APL   & Abductor pollicis longus        \\
BB    & Biceps brachii                  \\
DRUJ  & Distal radioulnar joint         \\
DTM   & Dart-throwing motion            \\
ECRB  & Extensor carpi radialis brevis  \\
ECRL  & Extensor carpi radialis longus  \\
ECU   & Extensor carpi ulnaris          \\
ED    & Extensor digitorum              \\
EDM   & Extensor digiti minimi          \\
EIP   & Extensor indicis proprius       \\
EPB   & Extensor pollicis brevis        \\
EPL   & Extensor pollicis longus        \\
FCR   & Flexor carpi radialis           \\
FCU   & Flexor carpi ulnaris            \\
FDP   & Flexor digitorum profundus      \\
FPL   & Flexor pollicis longus          \\
LT    & Lunotriquetral interosseous ligament \\
OP    & Opponens pollicis               \\
PQ    & Pronator quadratus              \\
PT    & Pronator teres                  \\
SL    & Scapholunate interosseous ligament \\
SP    & Supinator                       \\
TFCC  & Triangular fibrocartilage complex \\

\bottomrule
\end{tabular}
\end{table}
}

\newpage

\section{Introduction}
\label{sec:introduction}

\subsection{Background}

In order to manipulate a variety of tools and objects effectively in human-centered environments, humanoid robots must be able to adaptively adjust the stiffness of their wrist joints.
Unlike conventional single-axis robotic wrists, which have isotropic stiffness, the human wrist is believed to be able to adaptively adjust joint stiffness in both magnitude and direction based on task requirements.
This functionality is particularly evident along the dart-throwing motion (DTM),~\cite{palmer_functional_1985} a diagonal trajectory from radial extension to ulnar flexion that is preferentially used in activities of daily living.~\cite{anderton_movement_2022}
Prior studies have shown that passive {\color{red}human} wrist stiffness exhibits elliptical anisotropy with its minor axis aligned with the DTM direction,~\cite{formica_passive_2012} facilitating stable trajectory control during object manipulation.
Human joint stiffness is regulated by muscle co-contraction,~\cite{hogan_adaptive_1984,milner_adaptation_2002} and the human wrist can dynamically modulate the orientation of this stiffness ellipse accordingly.~\cite{zonnino_model-based_2018}
That is, low DTM-direction stiffness is maintained during grasping, whereas the stiffness ellipse can be rotated to increase DTM stiffness when resisting external disturbances.
Replicating this functionality in humanoid robots would significantly enhance manipulation stability; however, the morphological factors underlying this capability remain poorly understood.

\subsection{Related Work}

\begin{figure}[h]
  \centering
  \includegraphics[height=5.5cm]{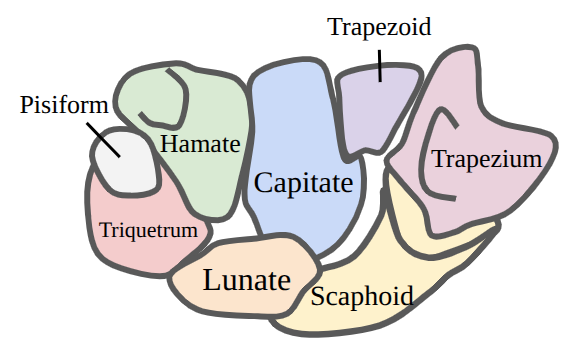}
  \caption{Carpal bone nomenclature}
  \label{fig:carpals}
\end{figure}

\subsubsection{Wrist Joint Stiffness in Humans}

Human wrist stiffness is anisotropic and can be represented by a stiffness ellipse in 2D.~\cite{crisco_mechanical_2011}
Measurements of passive wrist stiffness in subjects lightly grasping a handle consistently confirmed that the minor axis of the stiffness ellipse aligns with the DTM direction.~\cite{formica_passive_2012,durand_passive_2019,falzarano_evaluating_2021}
These studies suggested that the human wrist exhibits task-dependent anisotropic stiffness; however, since experimental manipulation of skeletal structure is infeasible in human subjects, the underlying morphological mechanisms remain unresolved.

In computational modeling, Zonnino et al.~\cite{zonnino_model-based_2018} used musculoskeletal simulation (OpenSim~{\color{red}\cite{delp_opensim_2007}}) to suggest that varying muscle co-contraction patterns can modulate wrist stiffness orientation.
However, this simulation modeled the wrist as a {\color{red}universal joint with elastic torque and damping, consolidating the eight carpal bones (Fig. \ref{fig:carpals}) into a single rigid body}.
The proximal carpal row acts as an intercalated segment~\cite{landsmeer_1961_study} possessing independent degrees of freedom and undergoing differential rotations under load;~\cite{garcia-elias_kinetic_1997,co_vivo_2005} consequently, the contribution of this carpal bone morphology to adaptive stiffness modulation could not be evaluated in that simulation.

\subsubsection{Biomimetic Wrist Robots}

In the field of biomimetic robotics, Kim et al.,~\cite{n_kim_bioinspired_2019} Hyeon et al.,~\cite{k_hyeon_lightweight_2023} and Chang et al.~\cite{m_h_chang_anthropomorphic_2022} developed prosthetic wrists, but either consolidated the carpal bones into a single unit or used {\color{red}simple geometric joint shapes that bear no resemblance to actual carpal bone morphology}.
He et al.\ placed a ball joint at the lunate position for compact, wide range of motion.~\cite{he_development_2025}
Design methods using ligament tensegrity structures have also been proposed,~\cite{sun_tensegrity_design_2020,hong_tensegrity_2019} and Hu et al.\ reproduced stiffness changes via pneumatic actuator jamming transition.~\cite{hu_biomimetic_2024}
Buchner et al.~\cite{buchner_replicating_2024} and Yang et al.~\cite{yang_enhancing_2024} developed highly biomimetic forearm robots incorporating ligaments and tendon routing, yet consolidated all eight carpal bones into a single rigid body, eliminating inter-carpal degrees of freedom.
The authors previously developed an anthropomimetic robot forearm faithfully reproducing the human wrist skeletal structure;~\cite{obata2023design} however, that prototype {\color{red}had its finger phalanges 3D-printed as single rigid pieces with fused joints, preventing finger flexion entirely, and additionally} had limitations in actuator controllability, making quantitative verification of stiffness modulation {\color{red}during grasping} difficult.

In summary, skeletal structure cannot be modified in human experiments, computational models and existing robots simplify the carpal bones, and the prior prototype had control constraints; consequently, the contribution of carpal bone morphology to adaptive stiffness modulation has not been quantitatively verified.

\subsection{Research Purpose}

The purpose of this study is to elucidate how skeletal morphology (carpal bone structure and joint surface geometry) and muscle arrangement contribute to adaptive stiffness modulation in the human wrist {\color{red}by using an anthropomimetic soft robotic forearm}.

To achieve this purpose, we extended the prior prototype~\cite{obata2023design,kumagai_improvement_2023,obata_development_2021} and developed an anthropomimetic soft robotic forearm equipped with eight independently movable carpal bones, 22 actuated muscles, {\color{red}fully articulated fingers,} and compliant dual-layer silicone fingertips.~\cite{kumagai_improvement_2023}
By conducting comparative experiments with different carpal skeletal configurations---which are infeasible with human subjects---we measure wrist stiffness under varying combinations of skeletal morphology and muscle activation patterns, thereby isolating and evaluating the morphological factors necessary for adaptive stiffness modulation.

\section{Materials and Methods}
\label{sec:methods}
\label{sec:prototype}

\subsection{System Overview}

\begin{table}[htbp]
  \centering
  \caption{Anthropomimetic robotic wrist specifications}
  \label{tab:spec}
  \begin{tabular}{ll}
    \toprule
    Component & Specification \\
    \midrule
    Actuators & 22 $\times$ Dynamixel XL330-M288-T \\
    {\color{red}Wrist muscles} & {\color{red}4 (FCR, FCU, ECRL/ECRB, ECU)} \\
    {\color{red}Forearm rotation muscles} & {\color{red}4 (SP, BB, PQ, PT)} \\
    {\color{red}Finger and thumb muscles} & {\color{red}14 (FDP $\times$4, FPL, EIP, EDM,} \\
                              & {\color{red}\phantom{14 (}ED, EPB, EPL, APL, APB, OP, AP)} \\
    Degree of freedom & 3 (Wrist) and 13 (Fingers) \\
    \bottomrule
  \end{tabular}

  \vspace{4pt}
  {\color{red}\footnotesize
  \begin{minipage}{0.85\linewidth}
  ECRL and ECRB are driven by a single shared actuator; FDP is independently actuated for each of the four fingers.
  \end{minipage}}
\end{table}

\begin{figure}[htbp]
  \centering
  \begin{subfigure}[b]{0.54\textwidth}
    \centering
    \includegraphics[width=\linewidth]{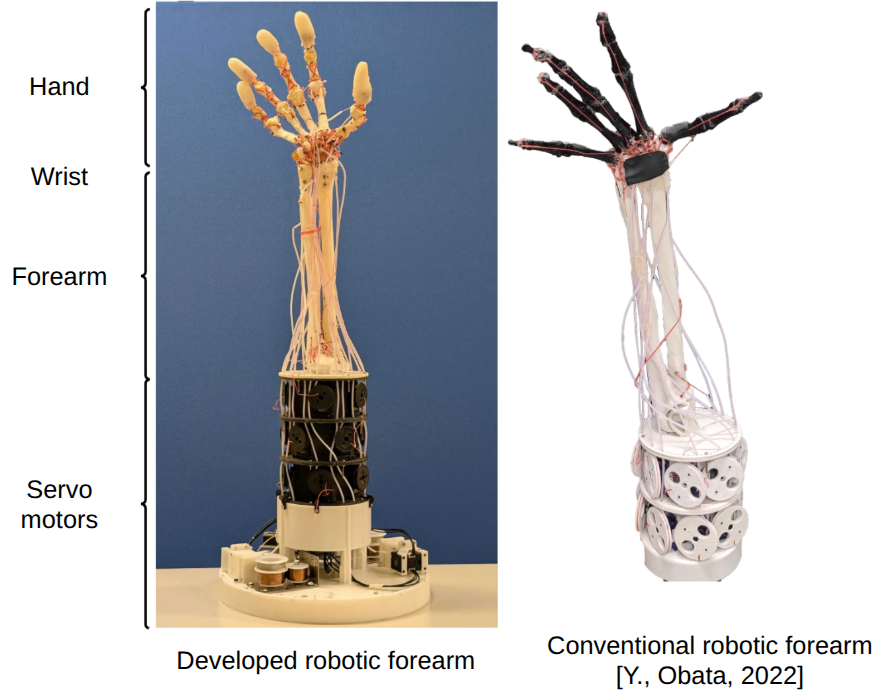}
    \caption{Integrated forearm robot system}
    \label{fig:prototype_overview}
  \end{subfigure}
  \hfill
  \begin{subfigure}[b]{0.44\textwidth}
    \centering
    \includegraphics[width=\linewidth]{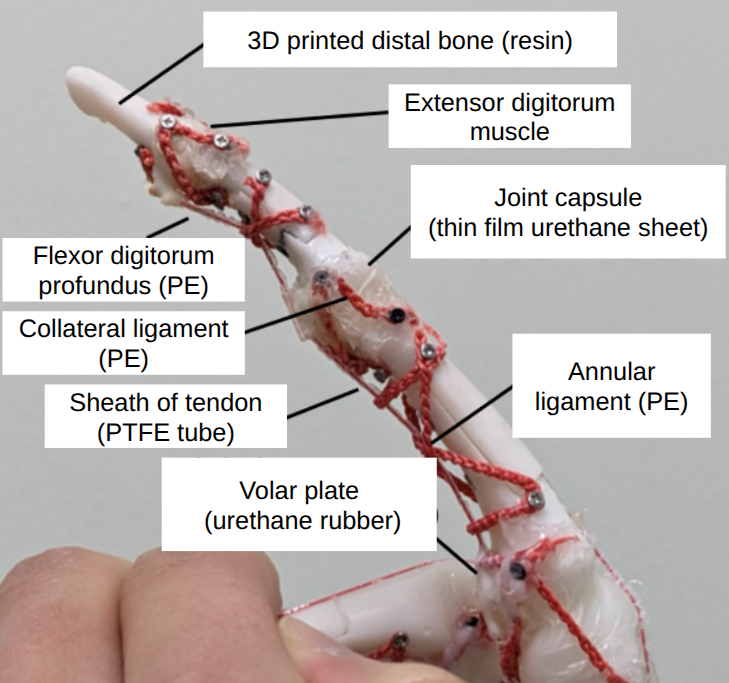}
    \caption{Fabricated anthropomimetic fingers}
    \label{fig:prototype_fingers}
  \end{subfigure}

  \vspace{2ex}
  \begin{subfigure}[b]{0.9\textwidth}
    \centering
    \includegraphics[width=\linewidth]{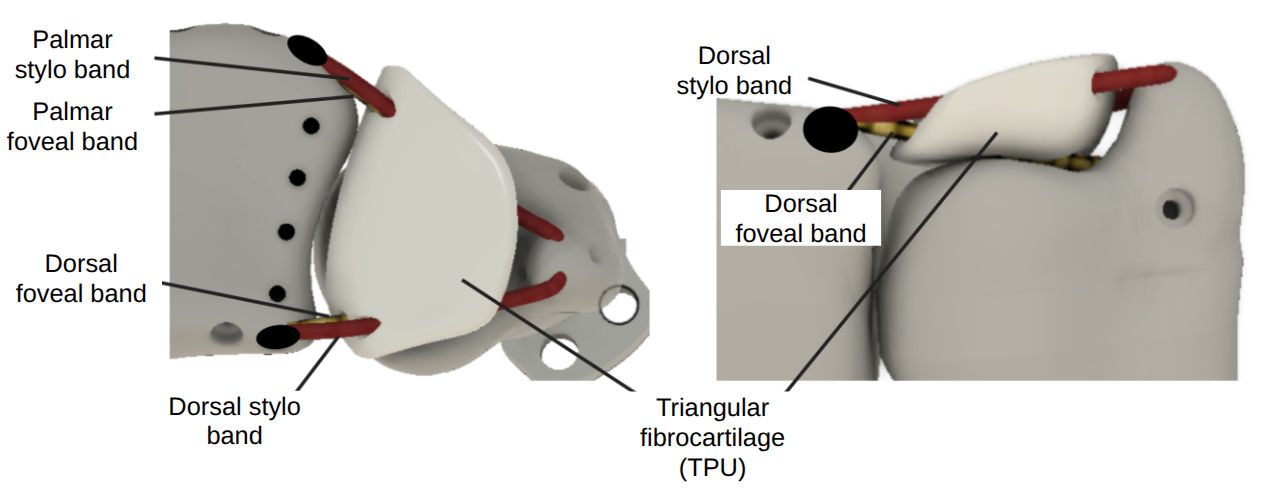}
    \caption{Structure of the triangular fibrocartilage complex (TFCC)}
    \label{fig:prototype_tfcc}
  \end{subfigure}
  \caption{Developed anthropomimetic soft robotic forearm. (a) Assembled system comprising 28 bones actuated by 22 servo motors. (b) Fabricated anthropomimetic fingers with dual-layer elastic fingertips. (c) TFCC, the primary stabilizer of the distal radioulnar joint, implemented with four radioulnar ligaments and a TPU-printed fibrocartilage disc.}
  \label{fig:prototype}
\end{figure}

\begin{figure}[htbp]
  \centering  
  \includegraphics[height=9cm]{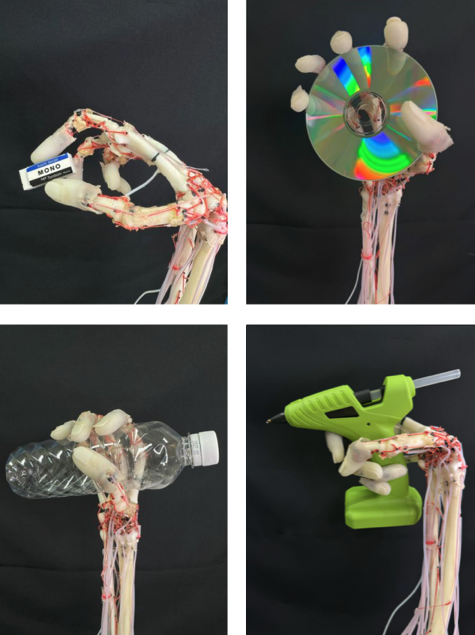}
  \caption{Examples of object grasping}
  \label{fig:grasp_examples}
\end{figure}

In this study, we reproduced the musculoskeletal structure of the human forearm and wrist based on anatomical knowledge to investigate the stiffness modulation mechanism of the human wrist.
The specifications of the robot{\color{red}ic} forearm are summarized in Table~\ref{tab:spec}.
The wrist joint configuration is based on our previous work.~\cite{obata2023design}
The developed robot forearm (Fig.~\ref{fig:prototype}a) comprises 28 bones (eight carpal, five metacarpal, fourteen finger phalanges, one radius, and one ulna) actuated by 22 servo motors (Dynamixel XL330-M288-T, Robotis Co.\ Ltd., Korea) with 12 mm radius pulley.
Since the robot forearm is equipped with wrist joints and five fingers, it can grasp various objects while actuating the wrist, as shown in Fig.~\ref{fig:grasp_examples}.
{\color{red}Supplementary Video~S1 shows the wrist actuated in each of its three degrees of freedom---flexion--extension, radial--ulnar deviation, and forearm pronation--supination---and wrist flexion performed while an object is held in the hand.}

\subsection{Wrist and Forearm Design}

\begin{figure}[htbp]
  \centering
  \includegraphics[height=9cm]{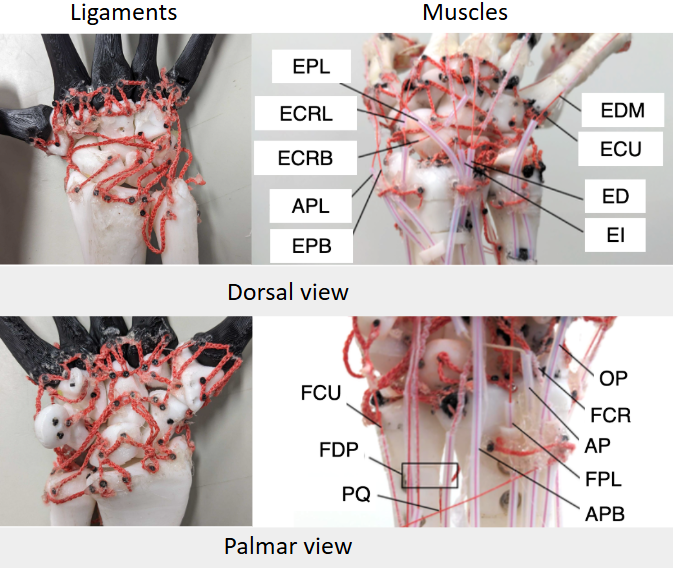}
  \caption{Arrangement of muscles and tendons (dorsal and palmar views)}
  \label{fig:muscle_arrangement}
\end{figure}

All bones were 3D-printed with photocurable resin from CT scan data of an adult male forearm,~\cite{BodyParts} and ligament attachment sites were designed based on anatomical knowledge.
Following our previous work,~\cite{obata2023design} we implemented the major carpal ligaments using chain-knitted polyethylene (PE) wire (diameter 0.23\,mm). The major ligaments include the palmar and dorsal radiocarpal ligaments, scapholunate (SL) and lunotriquetral (LT) interosseous ligaments, and palmar midcarpal ligaments.
Ligament lengths were set to achieve a passive range of motion matching the mean values of a Japanese population.~\cite{AIST} Dedicated assembly jigs were fabricated to ensure reproducible ligament lengths{\color{red}; the jigs and the full assembly procedure are documented in Supplementary Note~S1 (Figs.~S1--S5)}.

As shown in Fig.~\ref{fig:muscle_arrangement}, {\color{red}19 muscles (all except the three intrinsic thumb muscles)} were implemented using PE wire tendons (diameter 0.3\,mm) routed through PTFE tubes (outer diameter 2\,mm) serving as tendon sheaths.
Tendons originate from motor-driven pulleys mounted on a base plate and insert at anatomically appropriate positions on the metacarpals or phalanges.
Each motor employs current control (20--150\,mA depending on the muscle group) to achieve consistent muscle tension across experiments.

The triangular fibrocartilage complex (TFCC), the primary stabilizer of the distal radioulnar joint (DRUJ), was implemented with four radioulnar ligaments (dorsal/palmar $\times$ superficial/deep) and a TPU-printed triangular fibrocartilage disc (Fig.~\ref{fig:prototype}c).
The complete integrated system (Fig.~\ref{fig:prototype}a) {\color{red}shows the assembled forearm prototype with TFCC and finger assemblies integrated}.

\subsection{Finger Design and Soft Fingertips}

Following Xu et al.'s biomimetic approach,~\cite{xu_design_2016} we fabricated fingers (Fig.~\ref{fig:prototype}b).
For the thumb, {\color{red}three} intrinsic muscles (APB, OP, AP) were added to enable thumb opposition for grasping.

To enable wrist stiffness experiments during grasping, we developed biomimetic dual-layer elastic fingertips based on prior work.~\cite{kumagai_improvement_2023,obata_lifetech_2020,kumagai_asymmetric_2021} The fingertips consist of a hard outer layer (skin analog, Shore E30 silicone, 1\,mm thickness, Young's modulus 22.1\,kPa) and a soft inner layer (subcutaneous tissue analog, expanded silicone), with an integrated fingernail that resists dorsal deformation. {\color{red}The mould design and the casting procedure are given in Supplementary Note~S2 (Figs.~S6 and~S7).}

\subsection{Wrist Stiffness Measurement}

To verify our hypothesis that carpal bone morphology is necessary for adaptive stiffness modulability, we compared wrist joint stiffness across three skeletal types and four muscle activation patterns. To isolate skeletal morphology contributions from muscle force distribution effects, all motors were operated in current control mode generating identical muscle tension of 0.6\,N. This uniform activation approach allowed changes in wrist stiffness to be attributed solely to skeletal structure rather than muscle coordination strategies.

{\color{red}\subsubsection{Comparison Skeletons}}
\label{sec:comparison_skeletons}

\begin{figure}[htbp]
  \centering
  \includegraphics[height=4cm]{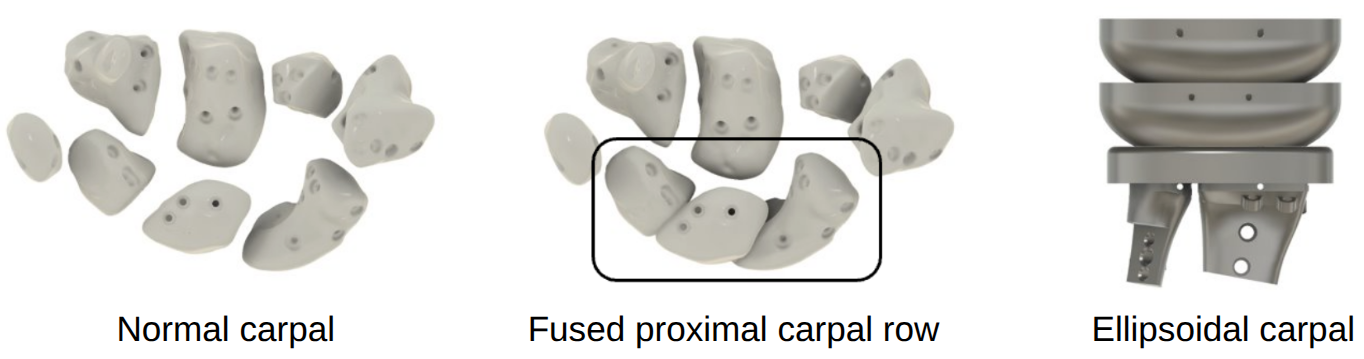}
  \caption{Three types of wrist skeletons compared}
  \label{fig:different_carpals}
\end{figure}

To investigate the contribution of skeletal morphology to adaptive stiffness modulation, we additionally fabricated two comparison skeletons and performed the same stiffness measurement (Fig.~\ref{fig:different_carpals}). The three skeletal types were: (1) \textit{Normal skeleton}---anatomically accurate human wrist skeleton with eight independently movable carpal bones; (2) \textit{Fused proximal row}---proximal carpal bones {\color{red}joined together in 3D CAD software}, eliminating inter-carpal degrees of freedom; and (3) \textit{{\color{red}Geometric} ellipsoidal skeleton}---redesigned from a bio-inspired prosthetic wrist~\cite{k_hyeon_lightweight_2023} to match human carpal bone size and radius-ulna morphology. The fused proximal row has only the midcarpal joint axis, matching the ellipsoidal skeleton's joint axis count. If wrist stiffness strongly depends on joint axis count, these two should show similar results; if joint surface morphology is critical, the fused proximal row should resemble the normal skeleton.

\subsubsection{Experimental Setup}

\begin{figure}[htbp]
  \centering
  \begin{minipage}[b]{0.53\linewidth}
    \centering
    \includegraphics[width=\linewidth]{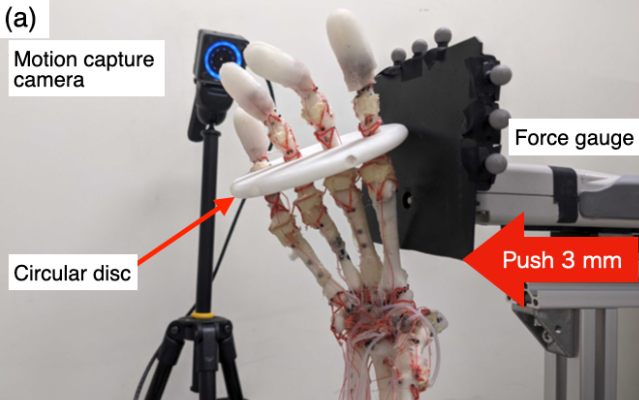}
  \end{minipage}
  \hfill
  \begin{minipage}[b]{0.43\linewidth}
    \centering
    \includegraphics[width=\linewidth]{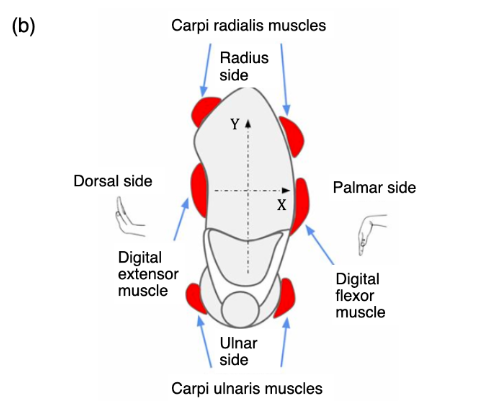}
  \end{minipage}

  \vspace{2ex}
  \begin{minipage}[b]{0.6\linewidth}
    \centering
    \includegraphics[width=\linewidth]{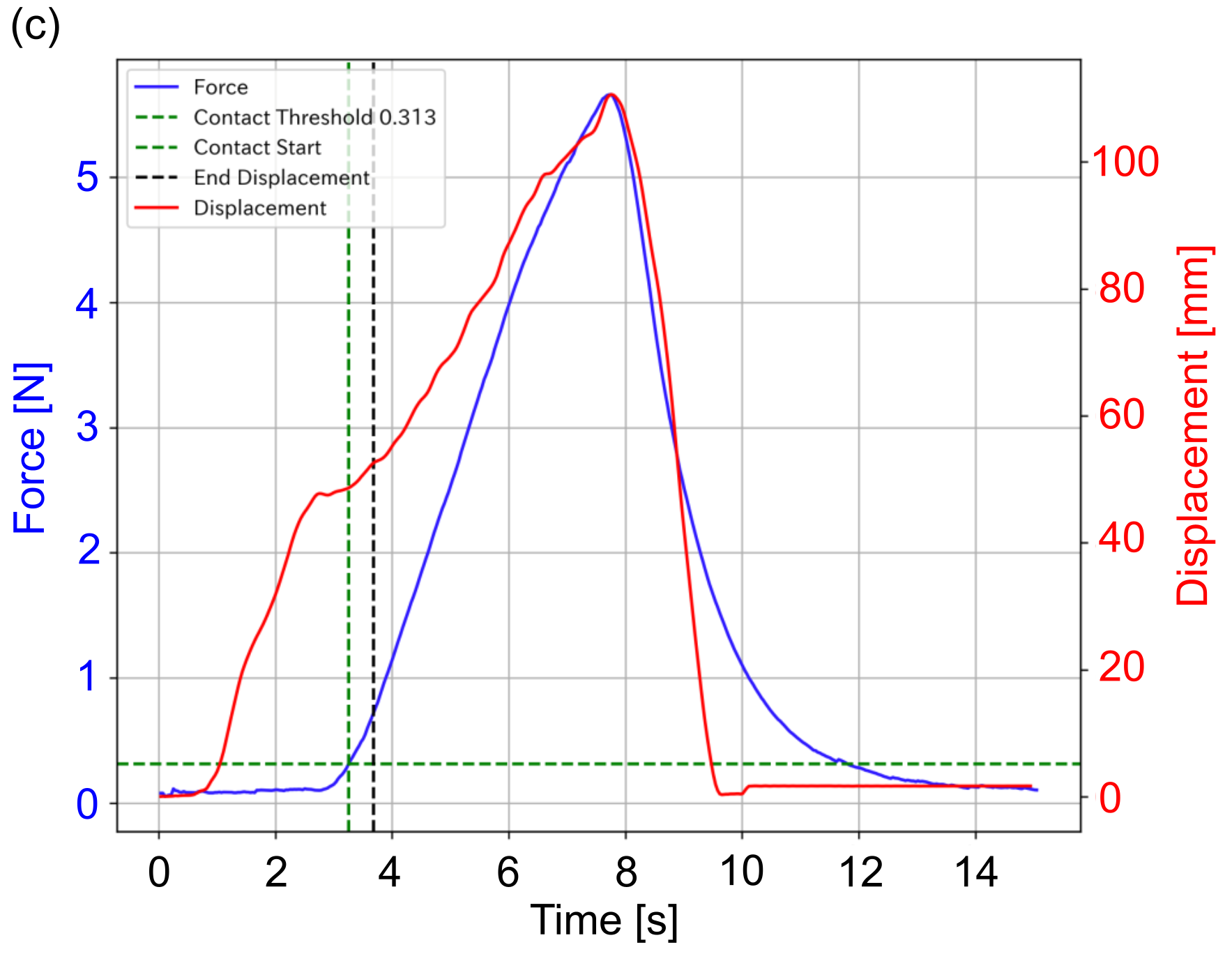}
  \end{minipage}
  \caption{Experimental setup for wrist stiffness measurement: {\color{red}(a) motion capture and force gauge configuration, (b) coordinate system and muscle layout, (c) representative force-displacement time series}}
  \label{fig:push_condition}
\end{figure}

Wrist stiffness was measured using a force gauge (ZTS-50N, IMADA Co.\ Ltd., Japan) mounted on a linear guide, pressed against a disc attached to the finger proximal phalanges to induce wrist deflection (Fig.~\ref{fig:push_condition}{\color{red}a}). Retroreflective markers on the force gauge plate were tracked by motion capture (OptiTrack Prime X13, NaturalPoint Inc., USA) at 100\,Hz, synchronized via trigger signal. We defined a coordinate system with origin at the midcarpal joint center, X-axis pointing palmarly (flexion direction), and Y-axis pointing radially (radial deviation direction){\color{red}, as shown in Fig.~\ref{fig:push_condition}b}.

\subsubsection{Experimental Procedure}

{\color{red}Four muscle activation patterns (relaxed, wrist muscles, finger muscles, and combined) were tested, as summarized in Table~\ref{tab:experiment_condition}.}

\begin{table}[htbp]
\centering
\caption{Muscle activation conditions}
\label{tab:experiment_condition}
\begin{tabular}{lll}
\toprule
Condition & Activated Muscles & Functional Scenario \\
\midrule
Relaxed & None & Passive wrist compliance \\
Wrist muscles & ECRL, ECRB, ECU, & Hand position stabilization \\
              & FCR, FCU & (typing, tapping) \\
Finger muscles & FDP, ED, EIP, EDM & Object grasping, \\
               &                   & inertial motion \\
Combined & Wrist + finger & Stabilized object grasp \\
\bottomrule
\end{tabular}
\end{table}

Force was applied from 12 directions at 30° intervals (0° to 330°). At each angle, the plate was pushed until 3\,mm displacement was achieved from contact. Eight trials were performed per condition per direction.

\subsubsection{Analysis Method}
\label{sec:analysis_method}

Wrist joint stiffness $K$ was calculated from {\color{red}the recorded force and displacement waveforms (Fig.~\ref{fig:push_condition}c)}:
\begin{equation}
  K = \frac{F_\mathrm{r}}{D}
\end{equation}
where $F_\mathrm{r}$ is resistance force and $D$ is displacement magnitude. The mean stiffness and standard deviation across eight trials were computed for each direction, excluding negative stiffness values (indicating structural coordinate misalignment) and values below 5\,N/m (attributable to incomplete contact detection). The 12-directional mean stiffness values were then transformed into a stiffness ellipse in the coordinate system. Stiffness ellipse parameters (major axis direction $\theta_\mathrm{max}$, minor axis direction $\theta_\mathrm{min}$, eccentricity $e$) were computed from constrained least-squares ellipse fitting.~\cite{fitzgibbon2002direct}

{\color{red}To test whether the reorientation of the maximum-stiffness direction ($\theta_\mathrm{max}$) between finger-muscle and combined-activation conditions reflected a genuine effect of muscle activation rather than trial-to-trial variability, we performed a stratified permutation test. Under the null hypothesis that $\theta_\mathrm{max}$ does not differ between the finger-muscle and combined-activation conditions, the per-trial stiffness values from these two conditions are interchangeable; we therefore pooled the per-trial stiffness values from the finger-muscle and combined-activation conditions and repeatedly shuffled the condition label within each of the 12 loading directions (preserving the direction-wise trial structure), each time recomputing $\theta_\mathrm{max}$ for the two relabeled groups and the circular difference between them. Repeating this 5000 times yielded a null distribution against which the observed difference could be compared.}

\subsection{Carpal Bone Motion Measurement}

\begin{figure}[htbp]
  \centering
  \includegraphics[height=7cm]{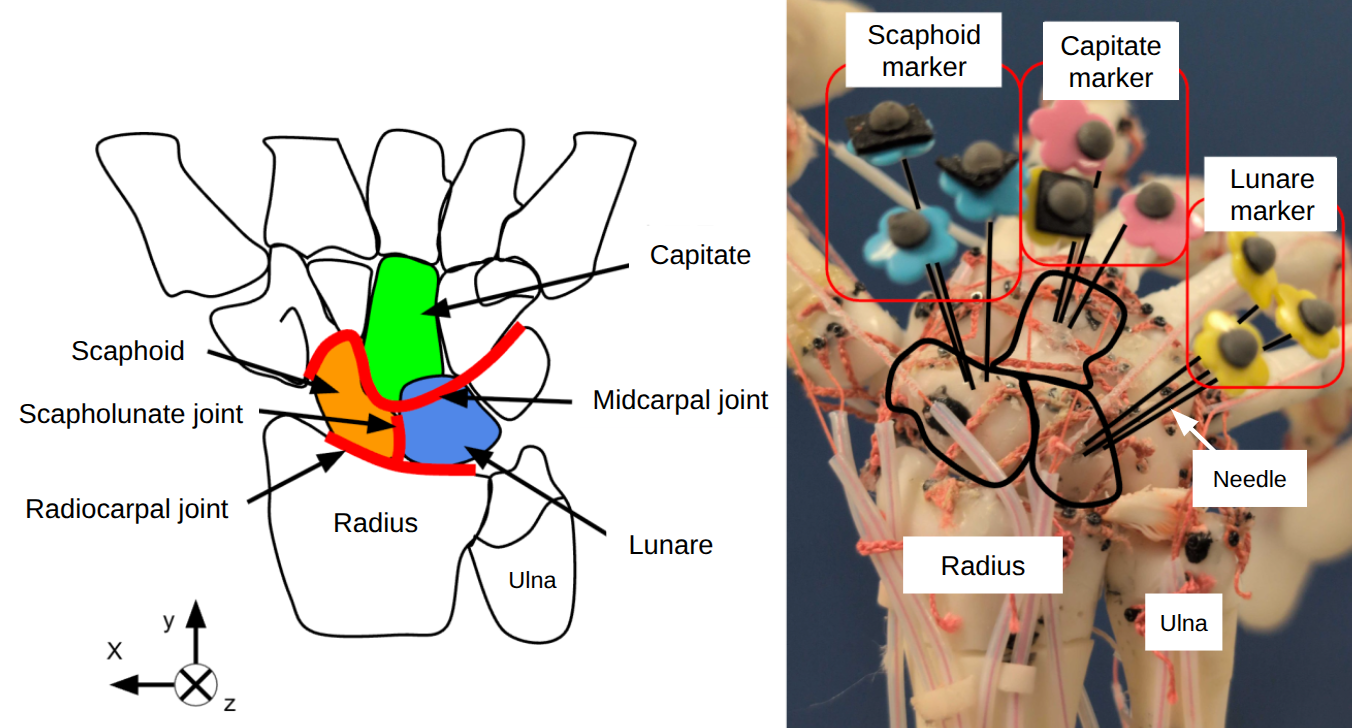}
  \caption{Marker pins inserted into scaphoid, capitate, and lunate, and simplified joint diagram}
  \label{fig:motion_measure_bones}
\end{figure}

To quantify the mechanical basis underlying the stiffness differences observed across skeletal types and muscle activation patterns, we measured relative motion between individual carpal bones during DTM-direction loading.

Markers placed directly on the carpal bones would be occluded by the ligament and tendon wires, and the rotational displacements between carpal bones are extremely small. We therefore inserted marker-tipped pins (2\,cm long) into the scaphoid, capitate, and lunate (Fig.~\ref{fig:motion_measure_bones}). Elevating markers above the bone surface amplifies their displacement, enabling reliable motion capture measurement. Loading was applied at 330°, the direction exhibiting the largest stiffness variation among conditions. The plate was slowly pushed while recording three-dimensional relative displacement and rotation of the three bones using the motion capture system. Twenty repetitions were performed per condition under three muscle activation patterns (wrist muscles, finger muscles, combined).

\section{Results}
\label{sec:results}

\subsection{Wrist Stiffness for Normal Skeleton}

\begin{figure}[htbp]
  \centering
  \begin{subfigure}[b]{0.52\textwidth}
    \centering
    \includegraphics[width=\linewidth]{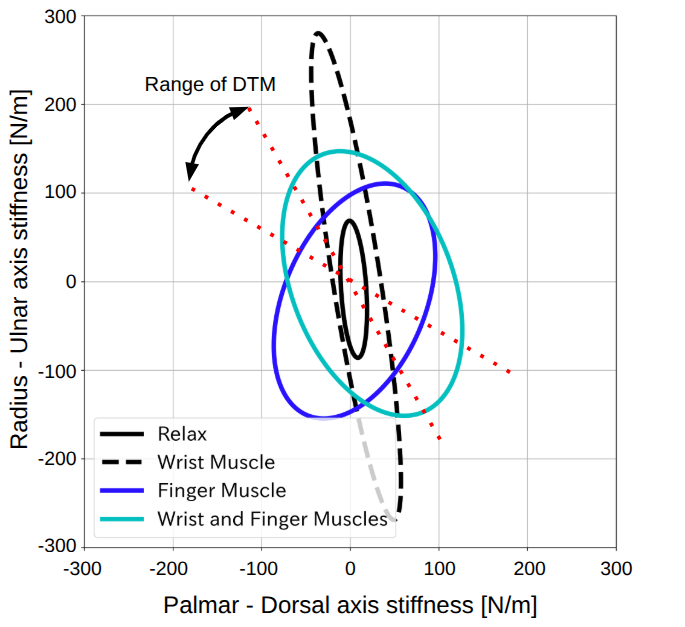}
    \caption{Normal carpal skeleton}
    \label{fig:ellipse_normal}
  \end{subfigure}

  \vspace{2ex}
  \begin{subfigure}[b]{0.48\textwidth}
    \centering
    \includegraphics[width=\linewidth]{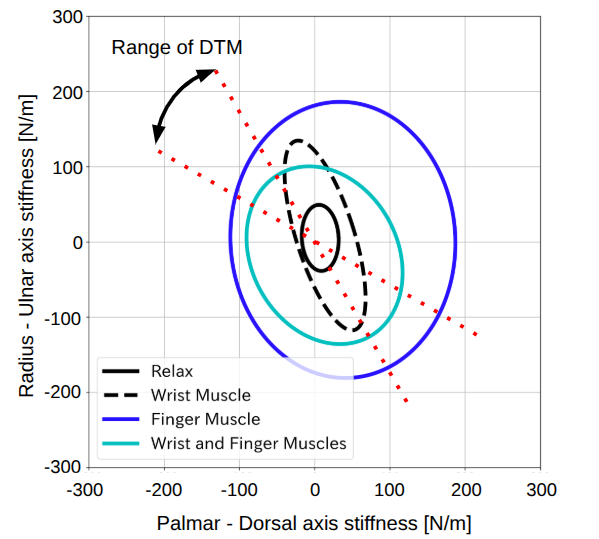}
    \caption{Fused proximal carpal row}
    \label{fig:ellipse_fused}
  \end{subfigure}
  \hfill
  \begin{subfigure}[b]{0.48\textwidth}
    \centering
    \includegraphics[width=\linewidth]{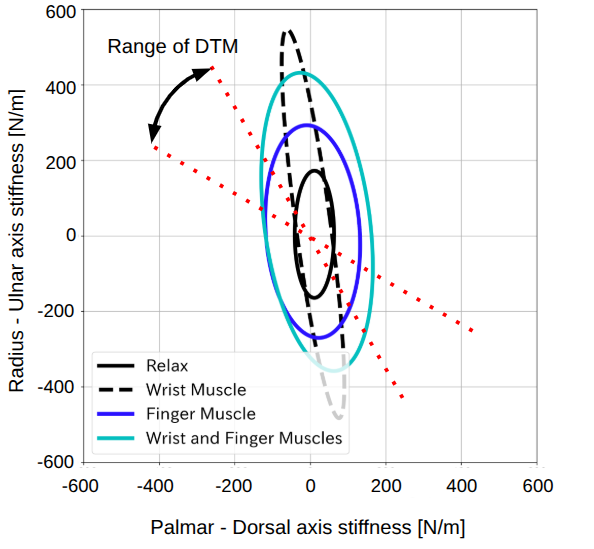}
    \caption{{\color{red}Geometric ellipsoidal} skeleton}
    \label{fig:ellipse_ellipsoidal}
  \end{subfigure}
  \caption{Wrist stiffness ellipses under the four muscle activation conditions for the three skeletal configurations. Panels (a) and (b) share the same axis range ($\pm 300$\,N/m), whereas panel (c) uses a twofold wider range ($\pm 600$\,N/m) to accommodate the higher absolute stiffness of the geometric ellipsoidal skeleton; ellipse sizes are therefore not directly comparable between (c) and the other two panels. Absolute stiffness values are listed in Tables~\ref{tab:ellipse_results}--\ref{tab:ellipse_ellipsoidal}.}
  \label{fig:stiffness_ellipses}
\end{figure}

\begin{table}[htbp]
\centering
\caption{Stiffness ellipse parameters for normal skeleton}
\label{tab:ellipse_results}
\begin{tabular}{lccccc}
\toprule
Condition & $\theta_\mathrm{min}$ & $\theta_\mathrm{max}$ & Eccentricity $e$ & $K_\mathrm{max}$ [N/m] & $K_\mathrm{min}$ [N/m] \\
\midrule
Relaxed                   & 182.4° & 275.9° & \textbf{0.816} &  77.4 & 14.2 \\
Wrist muscles activated   & 189.1° &  97.4° & 0.903 & 277.5 & 26.9 \\
Finger muscles activated  & \textbf{151.7°} & 254.0° & 0.431 & 140.7 & 80.1 \\
Combined activation       & 197.9° & \textbf{302.2°} & 0.418 & 156.1 & 90.8 \\
\bottomrule
\end{tabular}
\end{table}

Figure~\ref{fig:stiffness_ellipses}a and Table~\ref{tab:ellipse_results} show the stiffness ellipse results for the normal skeleton. Under finger muscle activation, the minor axis ($\theta_\mathrm{min} = 151.7^\circ$) nearly aligned with the DTM direction (120°--150°), indicating low resistance to motion in that direction. This result is consistent with human-subject experiments by Formica et al.~\cite{formica_passive_2012} and Durand et al.,~\cite{durand_passive_2019} demonstrating that the developed robotic wrist replicates key human wrist joint stiffness characteristics.

When both wrist and finger muscles were co-activated, the major axis ($\theta_\mathrm{max} = 302.2^\circ$) aligned with the DTM direction, indicating high resistance in that direction---in contrast to finger muscle activation alone, where instead the minor axis aligned with the DTM direction (Table~\ref{tab:ellipse_results}). This reorientation of the stiffness ellipse toward the DTM direction is consistent with the simulation predictions by Zonnino et al.,~\cite{zonnino_model-based_2018} and confirms real-space replication of human wrist adaptive stiffness modulability. {\color{red}To confirm that this reorientation was not attributable to trial-to-trial variability, we performed a stratified permutation test (Section~\ref{sec:analysis_method}): the observed shift in $\theta_\mathrm{max}$ between finger-muscle and combined activation ($\Delta\theta_\mathrm{max} = 48.2^\circ$) exceeded all but 18 of 5000 label-shuffled null rotations ($p = 0.0038$), indicating the reorientation is unlikely to arise from chance alone.}

In the relaxed condition, the stiffness ellipse was small with relatively high eccentricity ($e = 0.816$). Wrist muscle activation enlarged the ellipse while retaining similar orientation and eccentricity, indicating that wrist muscles increase overall stiffness without substantially reshaping the stiffness distribution.

\subsection{Comparison of Skeletal Configurations}

\begin{table}[htbp]
\centering
\caption{Stiffness ellipse parameters for fused proximal row skeleton}
\label{tab:ellipse_fixed}
\begin{tabular}{lccccc}
\toprule
Condition & $\theta_\mathrm{min}$ & $\theta_\mathrm{max}$ & Eccentricity $e$ & $K_\mathrm{max}$ [N/m] & $K_\mathrm{min}$ [N/m] \\
\midrule
Relaxed                   & 187.6° &  80.7° & 0.443 &  44.0 &  24.5 \\
Wrist muscles activated   & 197.9° & 100.1° & 0.707 & 131.6 &  38.6 \\
Finger muscles activated  & \textbf{183.4°} &  58.6° & \textbf{0.186} & 183.6 & 149.5 \\
Combined activation       & 187.4° & \textbf{301.0°} & 0.213 & 123.5 &  97.2 \\
\bottomrule
\end{tabular}
\end{table}

\begin{table}[htbp]
\centering
\caption{Stiffness ellipse parameters for geometric ellipsoidal skeleton}
\label{tab:ellipse_ellipsoidal}
\begin{tabular}{lccccc}
\toprule
Condition & $\theta_\mathrm{min}$ & $\theta_\mathrm{max}$ & Eccentricity $e$ & $K_\mathrm{max}$ [N/m] & $K_\mathrm{min}$ [N/m] \\
\midrule
Relaxed                   & 180.5° & \textbf{86.0°} & 0.691 & \textbf{168.5} &  52.1 \\
Wrist muscles activated   & 187.8° & \textbf{96.5°} & 0.911 & \textbf{518.3} &  46.1 \\
Finger muscles activated  & 185.2° & \textbf{92.3°} & 0.559 & \textbf{282.2} & 124.4 \\
Combined activation       & 189.3° & \textbf{94.0°} & 0.648 & \textbf{397.7} & 140.0 \\
\bottomrule
\end{tabular}
\end{table}

The fused proximal row skeleton showed reduced ellipse eccentricity compared to the normal skeleton, indicating reduced stiffness anisotropy (Fig.~\ref{fig:stiffness_ellipses}b and Table~\ref{tab:ellipse_fixed}). Particularly, finger muscle activation lost the low-DTM-direction stiffness characteristic observed in the normal skeleton ($\theta_\mathrm{min} = 183.4^\circ$), with eccentricity dropping sharply to $e = 0.186$. {\color{red}This drop reflects a marked increase in the minimum-axis stiffness, which lies near the palmar--dorsal direction ($\theta_\mathrm{min} \approx 180^\circ$): $K_\mathrm{min}$ rose from $80.1$\,N/m in the normal skeleton to $149.5$\,N/m in the fused skeleton under finger muscle activation (Tables~\ref{tab:ellipse_results} and~\ref{tab:ellipse_fixed}).} Under wrist muscle activation, the fused proximal row produced stiffness ellipses similar to those of the normal skeleton ($\theta_\mathrm{max} = 100.1^\circ$, close to the normal skeleton's $97.4^\circ$). Under combined activation, $\theta_\mathrm{max} = 301.0^\circ$ closely matched the DTM direction, again similar to the normal skeleton.

The {\color{red}geometric ellipsoidal} skeleton showed minimal ellipse orientation changes across all muscle activation conditions ($\theta_\mathrm{max} \approx 86^\circ$--$96^\circ$; Fig.~\ref{fig:stiffness_ellipses}c and Table~\ref{tab:ellipse_ellipsoidal}), with the stiffness ellipse orientation remaining nearly constant regardless of muscle activation pattern. {\color{red}The geometric ellipsoidal skeleton exhibited the highest absolute stiffness among the three types: its major-axis stiffness $K_\mathrm{max}$ exceeded that of the normal and fused skeletons in every activation condition (e.g., $282$--$518$\,N/m under wrist and finger activation, versus $140$--$278$\,N/m for the normal skeleton under the same two conditions; Tables~\ref{tab:ellipse_results}--\ref{tab:ellipse_ellipsoidal}), attributable to its large joint contact area.}

These results demonstrate that carpal bone morphology significantly impacts variable stiffness function. Normal carpal structure exhibited stiffness anisotropy and variability; fused carpal bones showed reduced anisotropy; and the {\color{red}geometric ellipsoidal} structure showed significantly decreased variability.

\subsection{Carpal Bone Motion}

\begin{figure}[p]
  \centering
  \begin{subfigure}[b]{\textwidth}
    \centering
    \includegraphics[width=\linewidth]{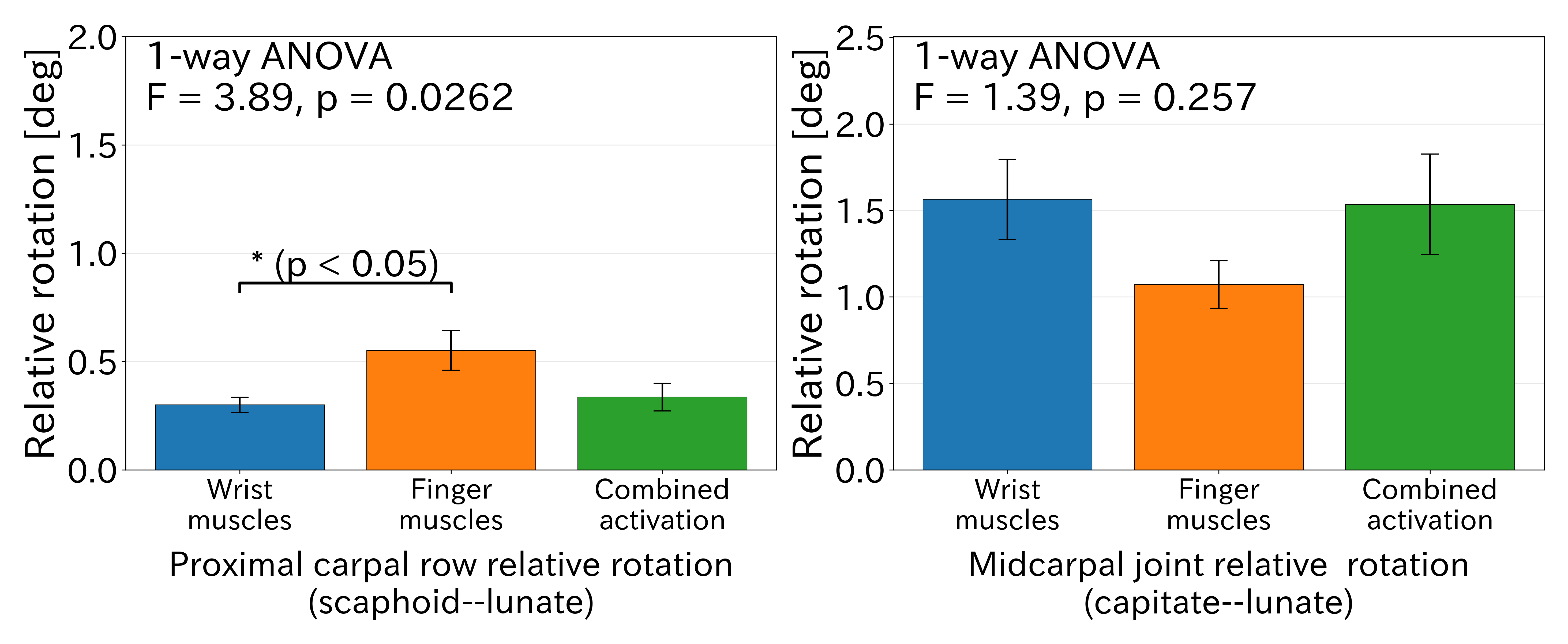}
    \caption{{\color{red}Relative rotation of scaphoid and capitate in lunate coordinate system}}
    \label{fig:motion_rotation}
  \end{subfigure}

  \vspace{1.5ex}
  \begin{subfigure}[b]{\textwidth}
    \centering
    \includegraphics[width=\linewidth]{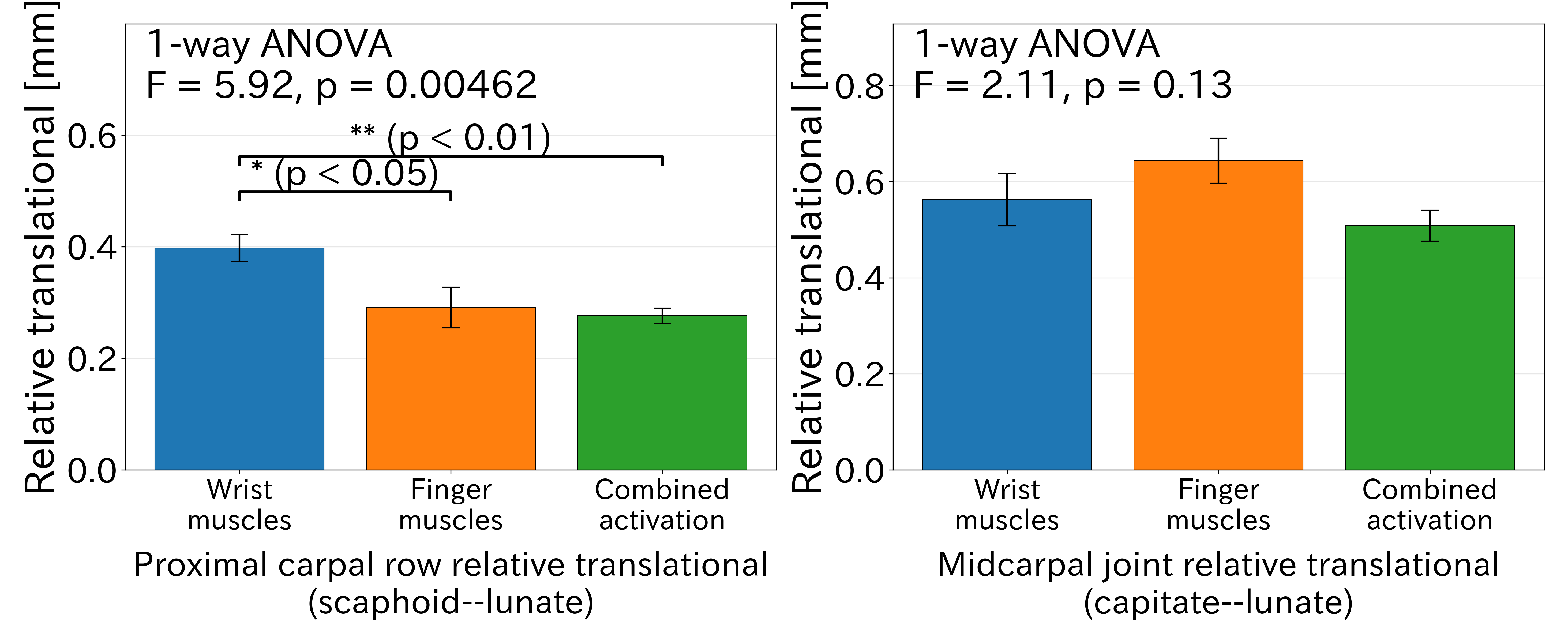}
    \caption{{\color{red}Relative translational displacement of scaphoid and capitate in lunate coordinate system}}
    \label{fig:motion_translation}
  \end{subfigure}

  \vspace{1.5ex}
  \begin{subfigure}[b]{0.5\textwidth}
    \centering
    \includegraphics[width=\linewidth]{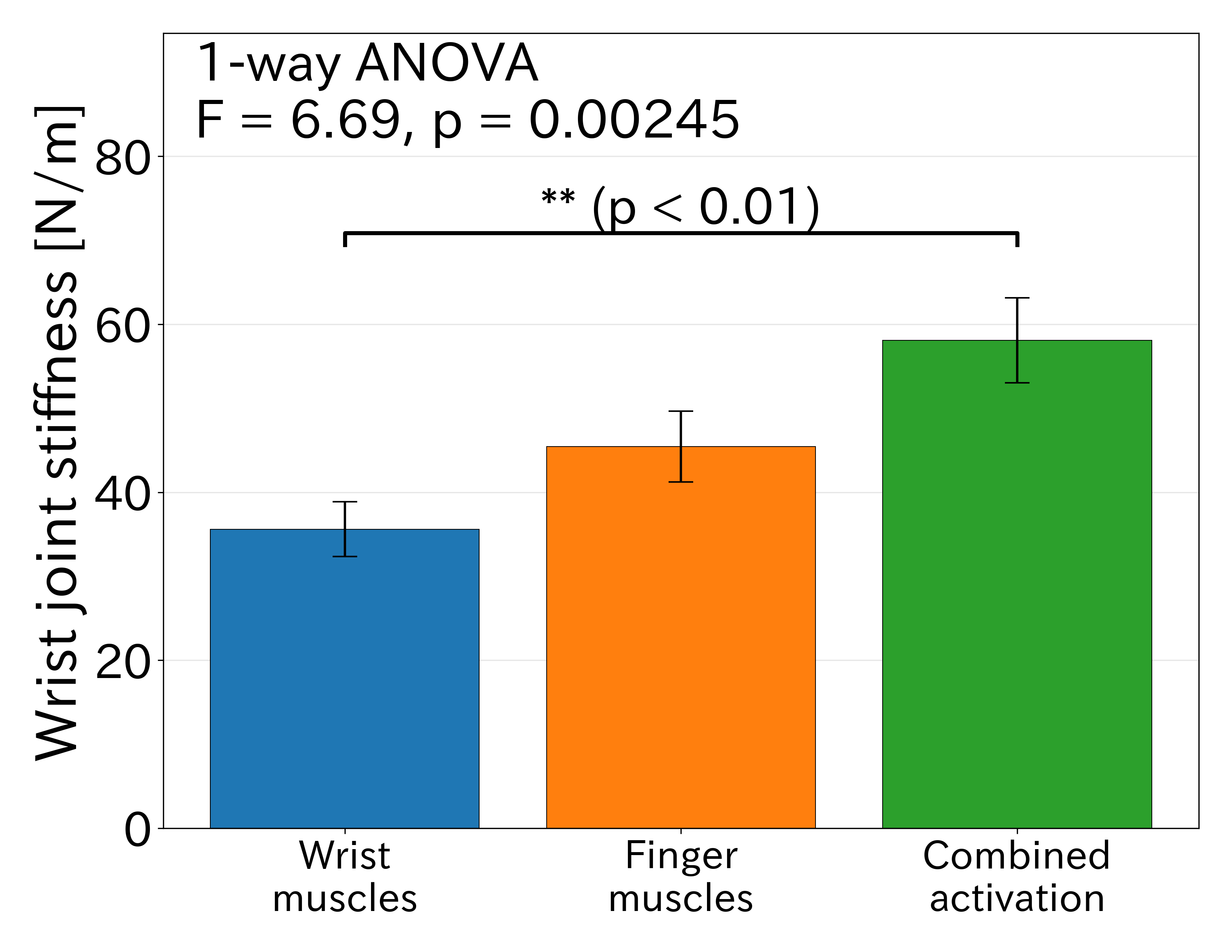}
    \caption{{\color{red}Wrist joint stiffness at 330°}}
    \label{fig:motion_stiffness}
  \end{subfigure}
  \caption{Carpal bone kinematics and wrist joint stiffness under the three muscle activation conditions (wrist muscles, finger muscles, and combined), measured during 330° loading.}
  \label{fig:carpal_motion}
\end{figure}

Figure~\ref{fig:carpal_motion}a and b show the three-dimensional relative rotation and translational displacement of the capitate and scaphoid in the lunate coordinate system. Figure~\ref{fig:carpal_motion}c shows the corresponding wrist joint stiffness.

{\color{red}One-way ANOVA showed a significant effect of muscle activation condition on wrist joint stiffness ($F=6.69$, $p=0.0025$), with Tukey HSD confirming a significant increase from wrist muscle activation to combined activation ($p=0.0016$). Proximal carpal row (scaphoid--lunate) relative rotation and relative translation also varied significantly with condition ($F=3.89$, $p=0.026$ and $F=5.92$, $p=0.0046$, respectively), whereas midcarpal joint (capitate--lunate) relative rotation and translation did not reach significance ($p=0.257$ and $p=0.13$).}

A negative correlation was observed between proximal carpal row relative rotation (scaphoid--lunate) and midcarpal joint relative rotation (capitate--lunate), {\color{red}suggestive of a compensatory relationship in which the two joints trade off rotation to maintain a constant total wrist rotation angle; since the midcarpal joint's condition-wise variation did not reach statistical significance, this compensatory pattern should be regarded as an observed trend rather than a confirmed effect}. For relative translational displacement, {\color{red}proximal row translation reached its highest value under wrist muscle activation, significantly exceeding both finger muscle activation ($p<0.05$) and combined activation ($p<0.01$; Tukey HSD).} The ranking of proximal row displacement across conditions was inversely related to the stiffness ranking---consistent with the expected inverse relationship between compliance and stiffness.

{\color{red}Proximal row relative rotation differed significantly between wrist and finger muscle activation (Tukey HSD, $p=0.033$): suppressed under wrist muscle activation and elevated under finger muscle activation, while midcarpal relative rotation showed the numerically opposite pattern (higher under wrist, lower under finger) without reaching significance. Under combined activation, both proximal row and midcarpal relative rotation reverted to values close to those observed under wrist muscle activation alone, consistent with both joints being co-stiffened; wrist joint stiffness reached its highest value under this condition, significantly exceeding wrist muscle activation alone (Tukey HSD, $p=0.0016$; Fig.~\ref{fig:carpal_motion}c)}.

\section{Discussion}
\label{sec:discussion}

The carpal bone motion measurements support the hypothesis derived from the skeletal comparison experiments, and confirm that muscle co-contraction switches the dominant joint axis between the proximal carpal row and midcarpal joint. Integrating these results, we discuss the mechanism of adaptive stiffness modulation and its morphological requirements.

\subsection{Validation and Consistency Across Skeletons}

The finger muscle activation condition in our experiment closely resembles prior human stiffness measurements, where subjects lightly grasp a handle while external torque flexes the wrist,~\cite{formica_passive_2012,durand_passive_2019} and our normal skeleton results are consistent with these measurements. Combined muscle activation's reorientation of the stiffness ellipse toward the DTM direction further corroborates the simulation predictions of Zonnino et al.~\cite{zonnino_model-based_2018}. Crucially, we achieved this validation through real-world mechanical implementation rather than computational simulation, demonstrating that the morphological features we incorporated are sufficient to produce human-like behavior.

A feature common to all three skeletal types was that the major axis of the stiffness ellipse was tilted toward the dorsal-radial direction, except under finger muscle activation in the normal and fused proximal row skeletons ($\theta_\mathrm{max}$ outside the $90^\circ$--$180^\circ$ range); the geometric ellipsoidal skeleton retained this tilt even under finger muscle activation ($\theta_\mathrm{max} = 92.3^\circ$), consistent with its minimal orientation change across all conditions. This orientation is consistent with the inclination angle of the radial articular surface, suggesting that the radial slope is a baseline determinant of wrist stiffness orientation.

All motors were operated with identical muscle tension (0.6\,N), unlike the simulation study by Zonnino et al.~\cite{zonnino_model-based_2018} which demonstrated that complex differential muscle tensions can modulate stiffness. Despite this uniform activation---without any differential force tuning---simple switching between finger and wrist muscle groups produced significant reorientation of the stiffness ellipse. This indicates that adaptive stiffness modulation is governed primarily by the morphological characteristics of skeletal geometry and muscle arrangement, rather than by the complexity of muscle input patterns.

\subsection{Mechanistic Role of Carpal Morphology}

\begin{figure}[htbp]
  \centering
  \includegraphics[height=8cm]{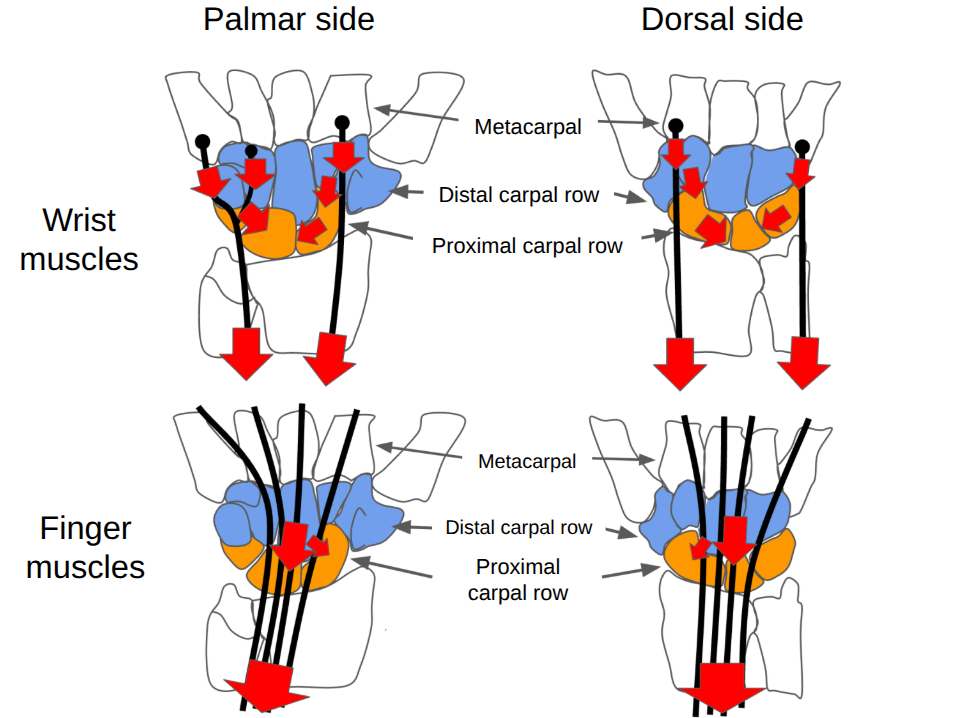}
  \caption{{\color{red}Compressive force pathways through the distal and proximal carpal rows under wrist-muscle activation (top; radial--ulnar tendon paths) and finger-muscle activation (bottom; palmar--dorsal tendon paths), shown for the palmar (left) and dorsal (right) views.}}
  \label{fig:carpal_force}
\end{figure}

Fusing the proximal carpal row eliminated individual bone degrees of freedom, causing joint incongruence that increased palmar-dorsal stiffness and reduced anisotropy, particularly under finger muscle activation. In contrast, the carpal bone motion experiment revealed that the normal proximal row redistributes load across activation conditions.

Wrist muscles, arranged radially and ulnarly, compress the scaphoid and lunate together (Fig.~\ref{fig:carpal_force}), {\color{red}significantly suppressing proximal row rotation relative to finger muscle activation, while significantly increasing proximal row translation relative to both other conditions; midcarpal rotation showed a non-significant upward trend---wrist-muscle coupling therefore selectively restricts rotation while leaving translation comparatively unconstrained.} Because rotation was restricted in both skeletons, the normal skeleton's stiffness ellipse under wrist muscle activation closely resembled the fused proximal row skeleton's; {\color{red}translation, however, set them apart, as the fused skeleton eliminates translation by construction while the normal row retains it.} Under finger muscle activation, tendons crossing the carpus palmarly and dorsally are hypothesized to compress the midcarpal joint (capitate--lunate interface) (Fig.~\ref{fig:carpal_force}), {\color{red}significantly increasing proximal row rotation} and transferring load through the compliant proximal row to produce low DTM stiffness. Under combined activation, {\color{red}both rotations reverted to near wrist-only values, consistent with co-stiffening as the radial--ulnar (wrist-muscle) and palmar--dorsal (finger-muscle) compression pathways acting together} (Fig.~\ref{fig:carpal_force}), yielding the highest overall stiffness and an ellipse reorientation toward the DTM direction.

This mechanical account rests primarily on the significant kinematic changes at the proximal carpal row; {\color{red}the accompanying midcarpal changes are directionally consistent but non-significant, and should be read as a supporting trend rather than a confirmed effect.} In the fused proximal row skeleton, the row cannot rotate, forcing all motion through the midcarpal joint; since low DTM stiffness under finger activation depends on load transfer through the compliant proximal row, the fused skeleton cannot replicate it.

Garcia-Elias et al.~\cite{garcia-elias_kinetic_1997} proposed that the proximal carpal row undergoes differential rotations under load to maintain force equilibrium, and Crisco et al.~\cite{co_vivo_2005} confirmed via CT-based motion tracking that each proximal carpal bone exhibits distinct rotational magnitudes during wrist motion. {\color{red}Our significant condition-dependent changes in proximal row relative rotation provide robotic experimental evidence supporting these biomechanical theories.}

Despite its highest absolute stiffness among the three skeletons (Table~\ref{tab:ellipse_ellipsoidal}), the {\color{red}geometric ellipsoidal} skeleton's uniform, symmetric joint surface distributes forces isotropically regardless of muscle activation direction, preventing the differential coupling needed for stiffness reorientation.

{\color{red}Returning to the two competing predictions from Section~\ref{sec:comparison_skeletons}, the fused proximal row and geometric ellipsoidal skeletons share the same joint axis count, yet their eccentricity trends diverged under finger and combined activation---ruling out axis count as the primary determinant of adaptive stiffness modulability. Supporting the joint-surface-morphology hypothesis instead, the fused row resembled the normal skeleton only when the normal proximal row is itself naturally consolidated into a single functional unit (wrist and combined activation), and diverged sharply under finger activation, where the normal row instead differentiates to redistribute load---a capacity the fused skeleton lacks by construction. The relevant requirement is therefore not joint surface shape per se, but the capacity for differential inter-carpal rotation it enables; permanent fusion eliminates this capacity regardless of surface shape.}

\subsection{Summary of Morphological Factors}

Based on the three experiments and their analysis, we identify the following primary morphological factors for human wrist adaptive stiffness modulability.
\begin{itemize}
  \item The proximal carpal row is subdivided into separate bones, each possessing flexion-extension and radial-ulnar degrees of freedom, enabling differential rotation under load
  \item Wrist muscles are arranged radially-ulnarly, while finger muscles cross the proximal carpal row palmarly-dorsally, creating orthogonal compression axes that produce condition-dependent inter-carpal coupling
  \item Proximal carpal bones have human-like asymmetric convex-concave morphology rather than simple geometric shapes, enabling direction-dependent force coupling
\end{itemize}

Applying these three design principles to humanoid robot wrists can realize human-like adaptive stiffness modulation with reduced control burden, as the morphology itself encodes the task-appropriate stiffness response.

{\color{red}\subsection{Limitations of the Present Study}}

{\color{red}This study has the following limitations. First, all wrist stiffness measurements were conducted under quasi-static loading conditions; dynamic behavior during manipulation tasks---such as rapid grasp adjustments or hammer swings---was not evaluated, and the morphological contributions identified here may behave differently under inertial or impact loading. Second, all results were obtained from a single physical prototype. While reproducibility within the build was systematically addressed through dedicated assembly jigs for ligament length control (Supplementary Note~S1) and current-controlled actuation for consistent muscle tension, evaluation of inter-build variability across replicate anthropomimetic forearms is reserved for future work.}

\section{Conclusion}
\label{sec:conclusion}

We developed an anthropomimetic soft robotic forearm with eight independently movable carpal bones and 22 actuated muscles, and experimentally investigated the effects of skeletal morphology on wrist stiffness modulation. The normal skeleton reproduced human-like adaptive stiffness modulation under uniform muscle activation, while the fused proximal row and {\color{red}geometric ellipsoidal} skeletons showed impaired modulability. Carpal bone motion analysis revealed that wrist and finger muscles produce opposing inter-carpal coupling patterns—statistically significant at the proximal carpal row and observed as a consistent, non-significant trend at the midcarpal joint—providing a mechanical explanation for the observed stiffness differences. These findings identified three primary morphological factors for adaptive stiffness modulability: independent proximal carpal bone degrees of freedom, orthogonal muscle arrangement across the carpal row, and asymmetric convex-concave joint surface morphology.

Future work will focus on investigating carpal bone behavior during dynamic manipulation tasks such as hammer swings, measuring wrist stiffness changes under diverse finger usage conditions including precision grasping and musical instrument playing, and developing simplified wrist structures incorporating these morphological principles for humanoid robots.

\section*{Author Contributions}

Yoshinobu Obata: Conceptualization, Methodology, Software, Investigation, Formal analysis, Data curation, Visualization, and Writing---original draft. Yinlai Jiang: Methodology, Supervision, Validation, and Writing---review and editing. Hiroshi Yokoi: Resources, Supervision, Validation, and Writing---review and editing. Shunta Togo: Conceptualization, Methodology, Funding acquisition, Project administration, Supervision, and Writing---review and editing. All authors reviewed and approved the final manuscript.

\section*{Statements and Declarations}

\subsection*{Ethical Considerations}
Ethical approval was not required because this study did not involve human participants, human data, human tissue, or animals.

\subsection*{Consent to Participate}
Not applicable.

\subsection*{Consent for Publication}
Not applicable.

\subsection*{Declaration of Conflicting Interest}
The authors declared no potential conflicts of interest with respect to the research, authorship, and/or publication of this article.

\subsection*{Funding Statement}
{\color{red}This work was supported in part by JSPS KAKENHI Grant Numbers JP26K00910 and JP23H00166.}

\subsection*{Data Availability}
{\color{red}The materials supporting this study are publicly available in the GitHub repository (\url{https://github.com/TogoLab/anthropomimetic-forearm-carpal-stiffness}). These materials include the 3D CAD and 3D-printable part files for the bones and the assembly jigs, the source code and data used for the statistical analyses, the Arduino/Teensy control source code, and Supplementary Video~S1, which shows wrist motion in three degrees of freedom and wrist flexion while grasping an object. Supplementary Notes~S1 and~S2, which document the jig-based assembly procedure and the fingertip casting procedure, are provided at the end of this article.}

\printbibliography[title=References]

\clearpage
\newrefsection
\setcounter{figure}{0}
\setcounter{table}{0}
\setcounter{section}{0}
\renewcommand{\thefigure}{S\arabic{figure}}
\renewcommand{\thetable}{S\arabic{table}}
\renewcommand{\thesection}{S\arabic{section}}
\renewcommand{\thesubsection}{S\arabic{section}.\arabic{subsection}}

\begin{center}
{\Large\bfseries Supplementary Material}\\[2ex]
{\large Anthropomimetic Soft Robotic Forearm with Independently Articulated\\
Carpal Bones Enabling Human-Like Adaptive Stiffness Modulability}\\[2ex]
Yoshinobu Obata, Yinlai Jiang, Hiroshi Yokoi, and Shunta Togo\\[1ex]
{\small Corresponding author: Yoshinobu Obata, \texttt{obata.yoshinobu@uec.ac.jp}}
\end{center}

\bigskip

\noindent\textbf{Contents}
\begin{itemize}[nosep,leftmargin=2em]
  \item Supplementary Note S1. Jig-based assembly of the skeleton and ligaments (Figs.~S1--S5)
  \item Supplementary Note S2. Fabrication of the dual-layer elastic fingertips (Figs.~S6--S7)
  \item Supplementary Video S1. Wrist motion in three degrees of freedom and wrist flexion while grasping an object (available in the GitHub repository listed under Data Availability)
\end{itemize}

\bigskip

\section{Jig-Based Assembly of the Skeleton and Ligaments}
\label{sec:supp_assembly}

This note documents the fabrication and assembly procedure of the anthropomimetic forearm, with emphasis on the dedicated jigs that were used to make the ligament arrangement and the ligament resting lengths reproducible between builds. Because the wrist stiffness measured in this study is governed by the carpal bone arrangement and by the length at which each ligament becomes taut, hand assembly without jigs would introduce build-to-build variability that could not be separated from the morphological effects under investigation. All jigs were designed in 3D CAD together with the bone models and were 3D-printed from the same photocurable resin; the CAD and printable files are included in the repository listed in the Data Availability statement of the main text.

\subsection{Bone Models and Ligament Attachment Holes}

\begin{figure}[htbp]
  \centering
  \includegraphics[width=0.85\linewidth]{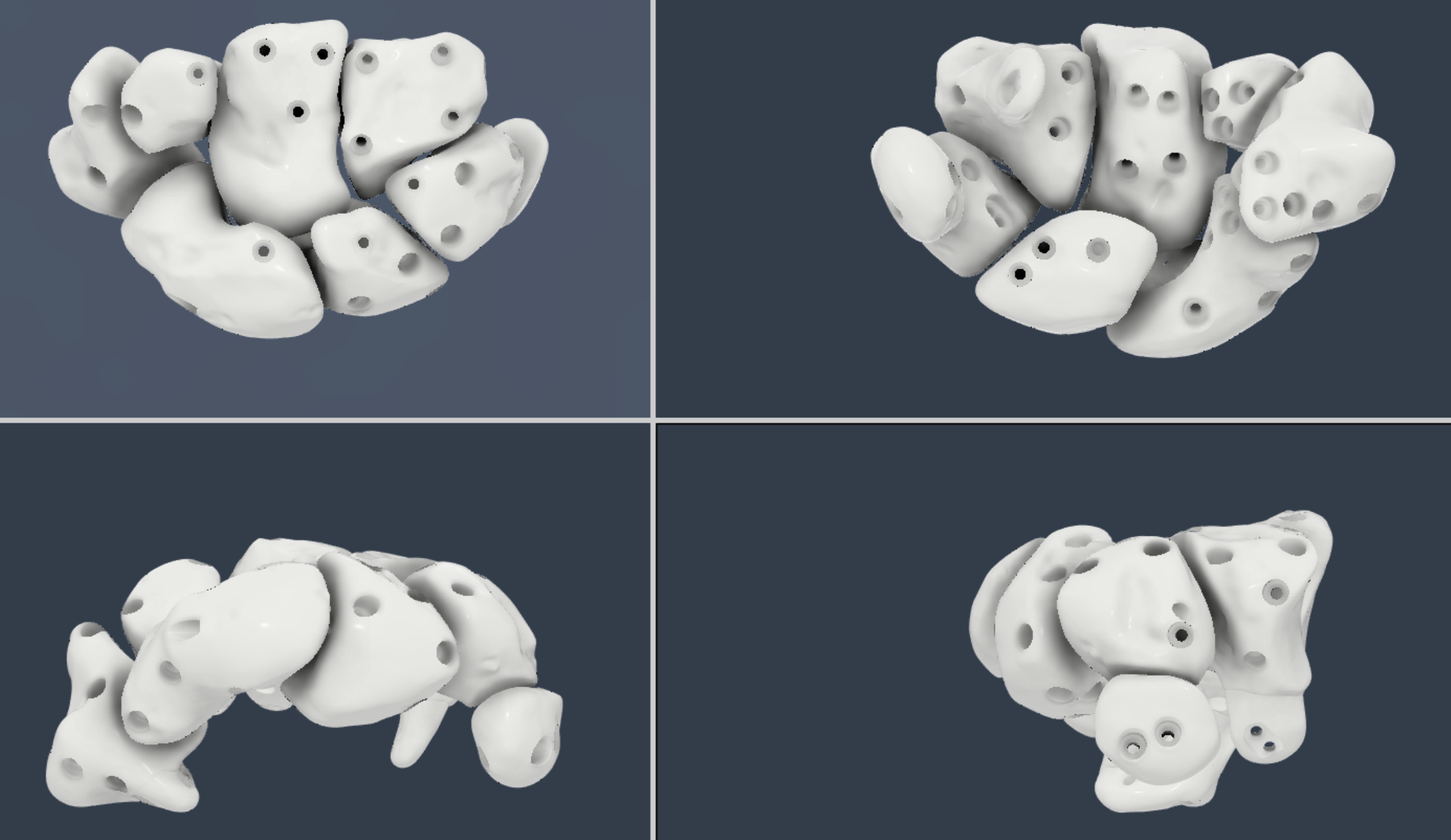}
  \caption{Ligament attachment holes designed on the carpal bones. The holes were positioned in 3D CAD at anatomically determined ligament attachment sites before printing, so that the same ligament arrangement is obtained every time the skeleton is rebuilt.}
  \label{fig:supp_holes}
\end{figure}

All 28 bones (eight carpal, five metacarpal, fourteen phalanges, radius, and ulna) were 3D-printed in photocurable resin from CT scan data of an adult male forearm.\cite{BodyParts} Ligament attachment holes were then added to each bone in 3D CAD at anatomically determined attachment sites (Fig.~\ref{fig:supp_holes}). Human carpal ligament attachments vary between individuals---the position of the dorsal transverse ligament in particular---so the most commonly reported configuration was used as the design reference. Two kinds of hole were needed. Most ligaments lie on the bone surface and are screwed to it, so a blind attachment hole at each end suffices. The interosseous ligaments, however, run between adjacent bones inside a carpal row and cannot be represented by a surface-mounted wire; for these, a through-hole was cut along the proximal row and along the distal row at the CAD stage, so that a single wire can be threaded through an entire row and anchored at each end (centre inset of Fig.~\ref{fig:supp_arrangement}). Placing both kinds of hole in the CAD model, rather than marking or drilling them on the printed bones by hand, is what makes the subsequent jig-based assembly repeatable: every rebuild starts from identical attachment geometry.

\subsection{Chain-Knitted Artificial Ligaments}

\begin{figure}[htbp]
  \centering
  \includegraphics[width=0.55\linewidth]{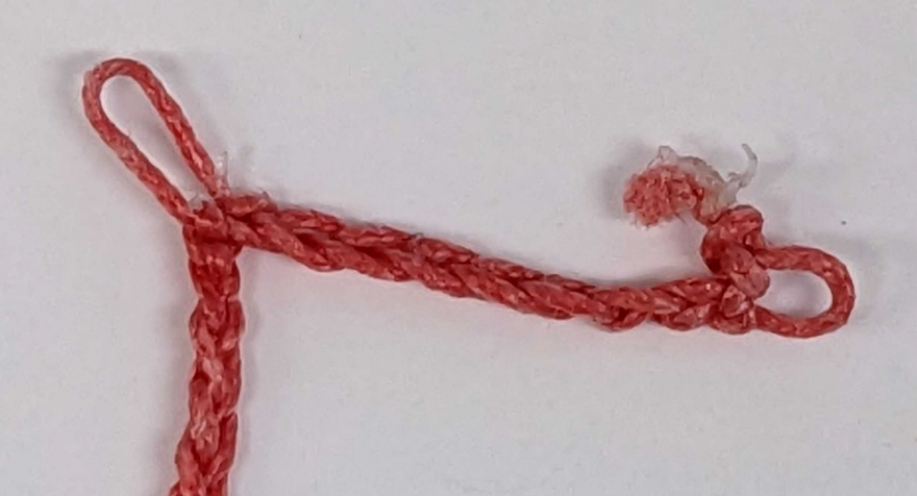}
  \caption{Artificial ligament chain-knitted from 0.23\,mm polyethylene wire. A screw passed through one of the mesh loops fixes the ligament to the bone and sets its resting length.}
  \label{fig:supp_pe}
\end{figure}

Human ligaments are stiff in tension---the elastic modulus of the male transverse carpal ligament is approximately 53\,MPa\cite{brett_quantification_2014}---which is orders of magnitude above that of rubber or silicone. Following the approach of Xu et~al.,\cite{xu_design_2016} artificial ligaments were therefore made by chain-knitting polyethylene (PE) wire of 0.23\,mm diameter (Fig.~\ref{fig:supp_pe}). The chain-knitted form has two practical advantages for assembly: a screw can be passed through any mesh loop, so a ligament can be fixed to a bone at an arbitrary point along its length, and the loop that is engaged by the screw defines the ligament resting length discretely and repeatably.

\subsection{Carpal Bone Holding Jig}

\begin{figure}[htbp]
  \centering
  \includegraphics[width=0.95\linewidth]{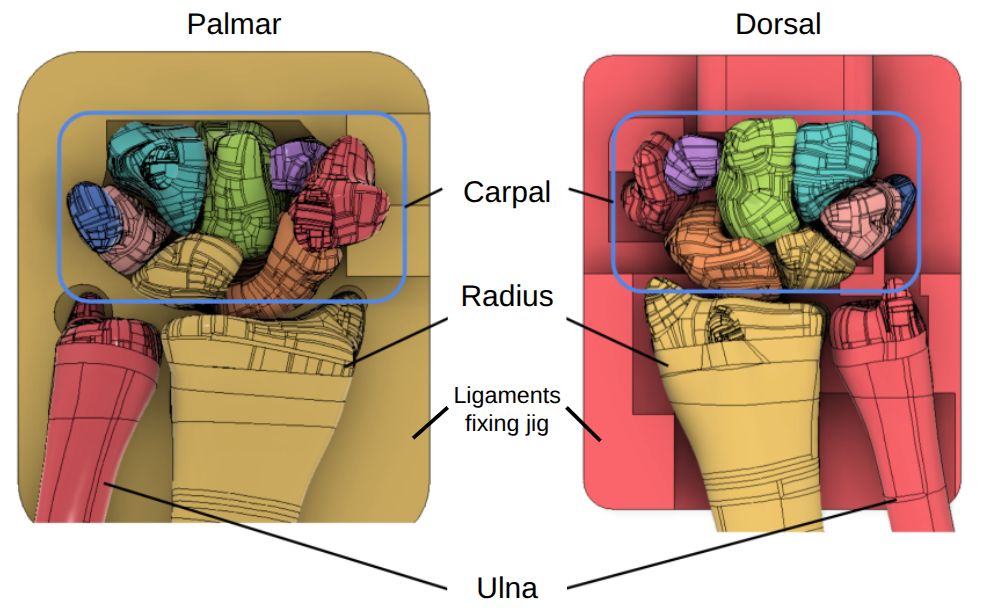}
  \caption{Carpal bone holding jigs used during ligament fixation. Each jig contains negative cavities of the carpal bones and of the distal radius and ulna, so that the bones are held in their anatomical arrangement while the ligaments on the opposite side are screwed in place.}
  \label{fig:supp_carpal_jig}
\end{figure}

The eight carpal bones are not rigidly connected to each other, so they cannot be held in their anatomical arrangement while ligaments are being screwed in place. A pair of holding jigs (Fig.~\ref{fig:supp_carpal_jig}) was therefore printed with negative cavities of the carpal bones, the distal radius, and the distal ulna, taken directly from the CT-derived CAD geometry. The palmar jig holds the skeleton while the dorsal ligaments are fixed, and the dorsal jig holds it while the palmar ligaments are fixed. With every bone seated in its cavity, the inter-carpal spacing and orientation at the moment of ligament fixation are set by the jig rather than by hand, which is the main source of reproducibility in the assembly.

\subsection{Ligament Length Setting Jig}

\begin{figure}[htbp]
  \centering
  \includegraphics[width=0.62\linewidth]{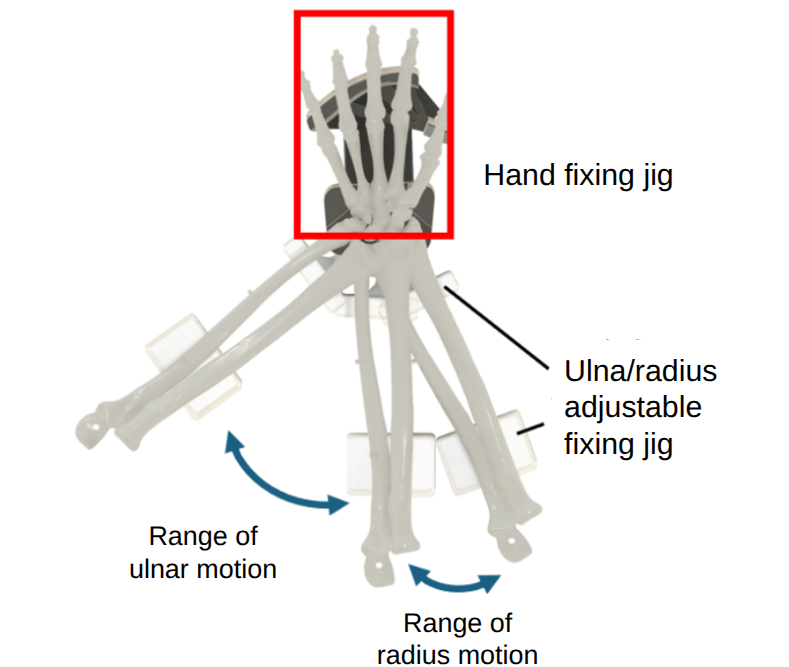}
  \caption{Jig for setting the ligament lengths. The fingers and carpal bones are held in a fixed holder while the radius and ulna are locked on movable arms at the target flexion--extension and radial--ulnar deviation angles taken from population mean values.}
  \label{fig:supp_radius_jig}
\end{figure}

\begin{figure}[htbp]
  \centering
  \includegraphics[width=0.7\linewidth]{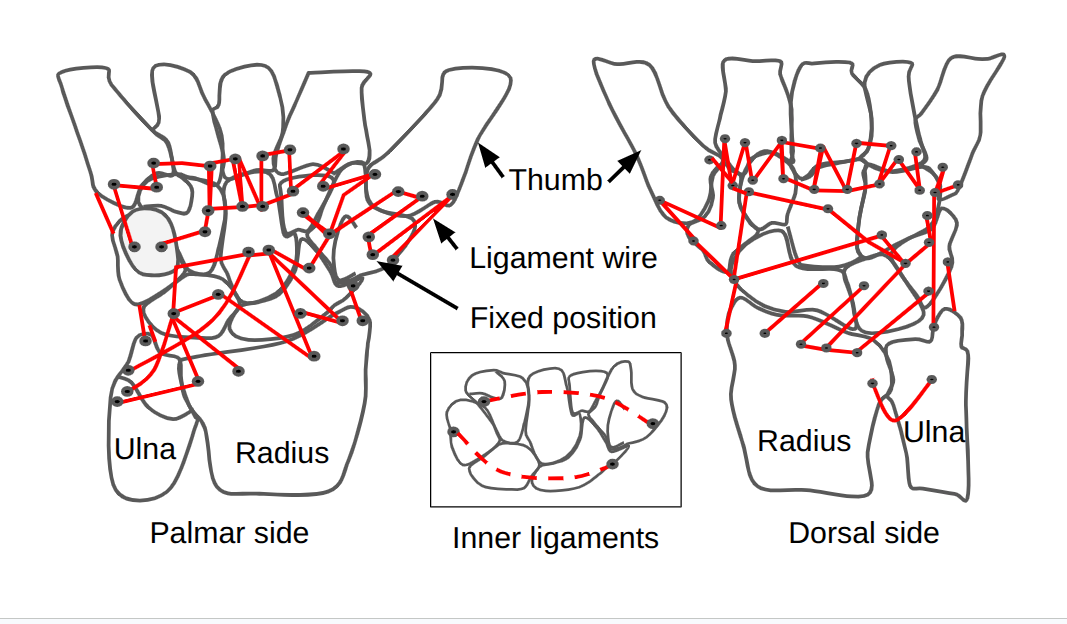}
  \caption{Artificial carpal ligament arrangement realised by the jig-based assembly, viewed from the palmar (left) and dorsal (right) sides. Bone outlines are traced from the CT-derived CAD model; red lines denote the chain-knitted polyethylene wire ligaments and the dots mark the fixed positions, i.e.\ the attachment holes at which each wire is screwed to the bone, which set its resting length. The centre inset shows the interosseous ligaments, which run between adjacent bones inside the proximal and the distal carpal row and therefore cannot be reproduced by a wire laid on the bone surface: a through-hole was drilled along each row and one wire was threaded through it and anchored at a fixed position at either end. Dashed segments indicate wire running inside the bones.}
  \label{fig:supp_arrangement}
\end{figure}

Ligament length directly determines the passive range of motion (ROM) of the wrist, and therefore the passive restoring torque that contributes to the measured stiffness. The target passive ROM was taken from mean values for a Japanese population.\cite{AIST} A second jig (Fig.~\ref{fig:supp_radius_jig}) holds the fingers and carpal bones in a fixed holder while the radius and ulna are carried on movable arms that can be locked at prescribed flexion--extension and radial--ulnar deviation angles. The skeleton was locked at the target maximum angle for each direction, and each ligament was then screwed at the mesh loop at which it just became taut. Repeating this for every ligament makes the passive ROM a designed quantity rather than an outcome of manual adjustment. The resulting arrangement is shown in Fig.~\ref{fig:supp_arrangement}: the implemented ligaments are the palmar and dorsal radiocarpal ligaments, the scapholunate and lunotriquetral interosseous ligaments, and the palmar midcarpal ligaments, reproducing the corresponding human carpal ligament arrangement on both the palmar and dorsal sides.

\subsection{Finger Joints}

The same principle was applied to the fingers. Each interphalangeal and metacarpophalangeal joint was assembled by wrapping an elastic urethane film joint capsule around the joint, bonding its ends, and then fixing the chain-knitted collateral ligaments with screws. During collateral ligament fixation, the joint was held by a joint-angle jig at the mean joint angle of a Japanese population, so that the ligaments become taut at the intended joint angle in every finger.

\subsection{Assembly Sequence}

The full sequence, in the order in which the jigs are used, is:
\begin{enumerate}[nosep]
  \item Print all bones with the CAD-defined ligament attachment holes (Fig.~\ref{fig:supp_holes}).
  \item Chain-knit the PE wire ligaments (Fig.~\ref{fig:supp_pe}).
  \item Seat the carpal bones, radius, and ulna in the palmar holding jig and fix the dorsal ligaments; repeat with the dorsal jig for the palmar ligaments (Fig.~\ref{fig:supp_carpal_jig}).
  \item Transfer the skeleton to the ligament length setting jig, lock it at each target maximum angle, and fix each ligament at the loop at which it just becomes taut (Fig.~\ref{fig:supp_radius_jig}).
  \item Assemble the finger joints with the joint-angle jig, then attach the fingers to the metacarpals.
  \item Route the tendons through their PTFE sheaths and connect them to the actuator pulleys.
\end{enumerate}

\section{Fabrication of the Dual-Layer Elastic Fingertips}
\label{sec:supp_fingertip}

\begin{figure}[htbp]
  \centering
  \includegraphics[width=0.62\linewidth]{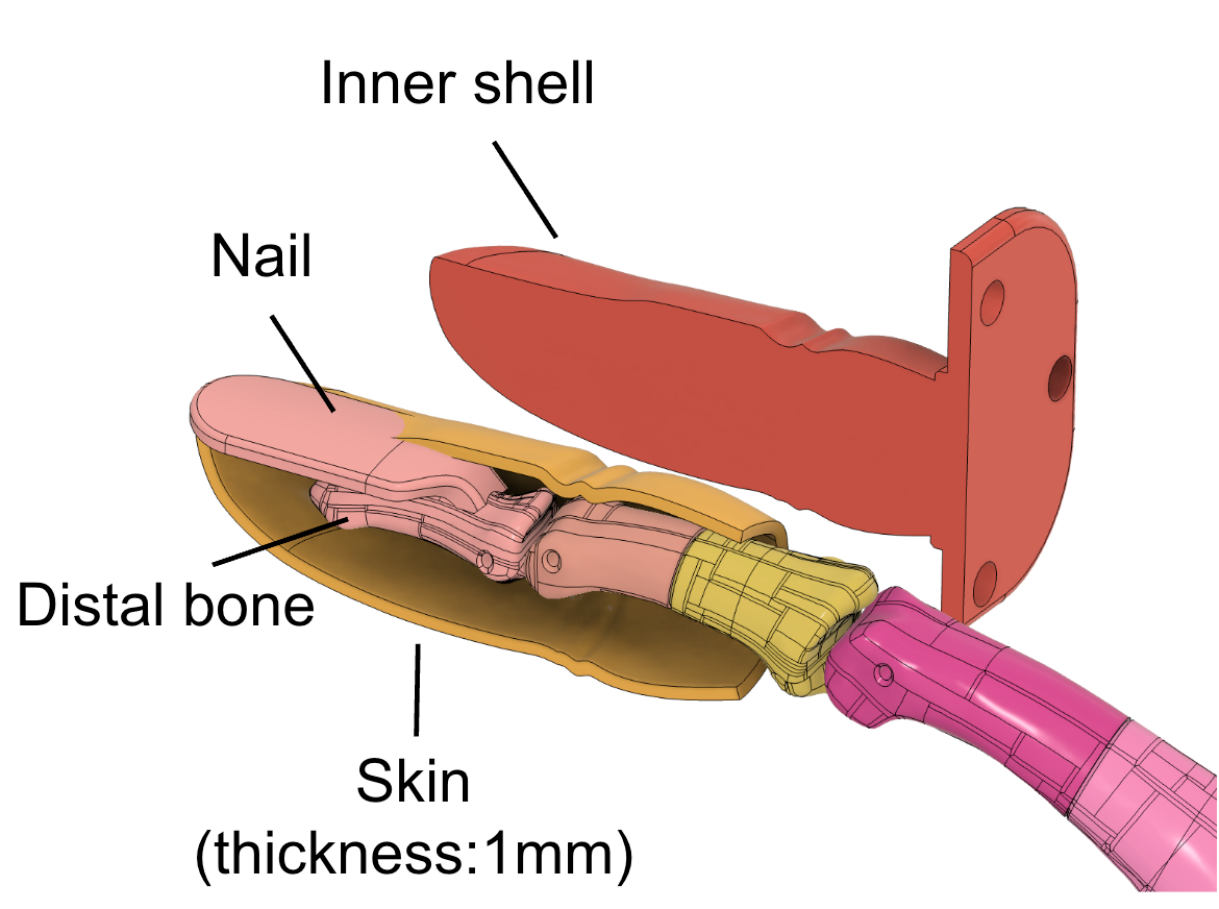}
  \caption{CAD design of the fingertip. The distal phalanx with the integrated fingernail is enclosed by a 1\,mm thick skin layer; the inner shell defines the cavity that is later filled with the expanded silicone subcutaneous layer.}
  \label{fig:supp_skin_design}
\end{figure}

\begin{figure}[htbp]
  \centering
  \includegraphics[width=0.52\linewidth]{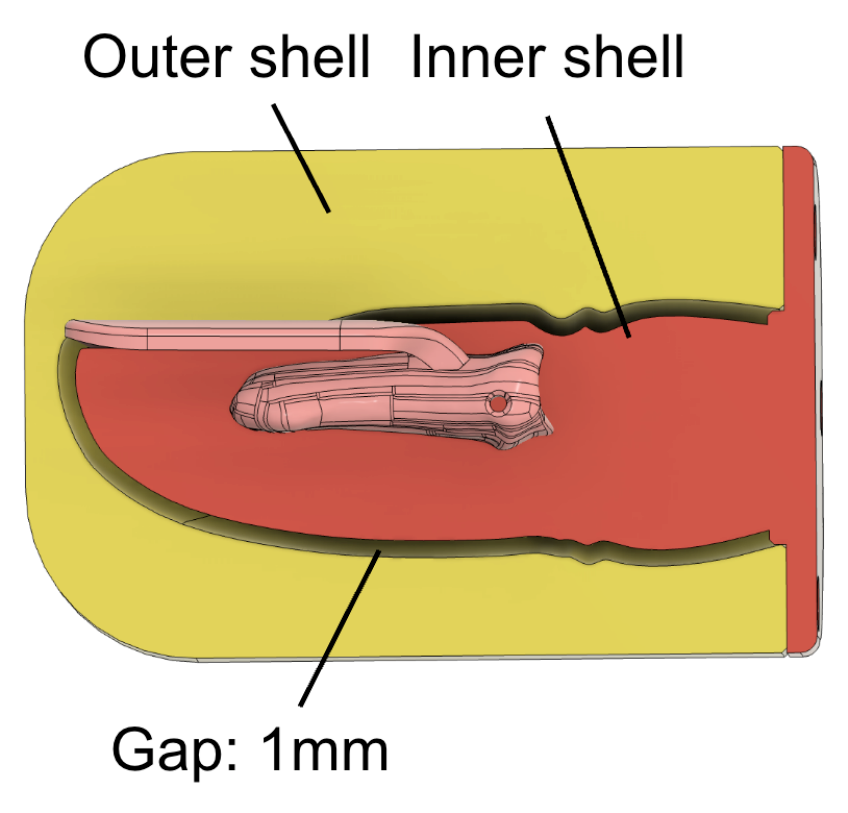}
  \caption{Cross-section of the two-part mould. The 1\,mm gap between the outer shell and the inner shell sets the thickness of the cast skin layer.}
  \label{fig:supp_skin_mold}
\end{figure}

The compliant fingertips used in the grasping conditions consist of a hard outer skin layer and a soft inner subcutaneous layer, cast in a two-part mould around the printed distal phalanx. The fingertip geometry (Fig.~\ref{fig:supp_skin_design}) is derived from the same CT-based finger CAD model: the distal phalanx with its integrated fingernail is enclosed by a 1\,mm thick skin layer, and an inner shell defines the cavity that is later filled with the subcutaneous layer. The mould is a two-part assembly in which the outer shell is offset from the inner shell by 1\,mm, so that the gap between them sets the cast skin thickness (Fig.~\ref{fig:supp_skin_mold}). Both mould halves were 3D-printed in photocurable resin.

The casting procedure is:
\begin{enumerate}[nosep]
  \item Apply a fluorine-based release agent to the outer and inner shells and assemble the mould.
  \item Pour the mixed two-part Shore~E30 silicone (TSG-E30, TANAC Inc.) into the 1\,mm gap.
  \item After curing, demould the skin under running water; this completes the outer skin layer.
  \item Insert the assembled finger skeleton, with its 3D-printed distal phalanx and integrated fingernail, into the skin.
  \item Pour the expanded silicone (TSG-SPO, TANAC Inc.) that forms the subcutaneous layer into the remaining cavity.
  \item After curing, bond the fingernail to the skin with adhesive.
\end{enumerate}

\noindent Because the silicone surface is tacky, powder was applied to the finished fingertip so that the surface condition was consistent across trials. The fingertip radius was set to 7.5\,mm, based on the mean distal phalanx width of Japanese adult males.\cite{AIST}

\printbibliography[title=Supplementary References]

\end{document}